\documentclass[11pt]{article}

\usepackage[final]{meta/acl}

\usepackage{times}
\usepackage{latexsym}

\usepackage[T1]{fontenc}

\usepackage[utf8]{inputenc}

\usepackage{microtype}

\usepackage{inconsolata}

\usepackage{graphicx}

\usepackage{bm}
\usepackage{amssymb}
\usepackage{amsmath}
\usepackage{enumitem}
\usepackage{tcolorbox}
\usepackage{caption,graphics,epstopdf,wrapfig}
\usepackage{threeparttable,tabularx,booktabs,multirow,multicol,verbatim}
\usepackage{lipsum}
\usepackage{listings}
\usepackage{subcaption}
\usepackage{makecell}

\definecolor{dark_red}{RGB}{189, 74, 87}
\newcommand{\cb}[1]{{\color{dark_red}{\textbf{#1}}}}

\newcommand{\secref}[1]{\hyperref[#1]{\S\ref*{#1}}}

\DeclareMathOperator*{\EV}{\mathbb{E}}

\title{Identifying and Transferring Reasoning-Critical Neurons: Improving LLM Inference Reliability via Activation Steering}

\author{
Fangan Dong$^{1}$, Zuming Yan$^{1}$, Xuri Ge$^{1}$, Zhiwei Xu$^{1}$, 
Mengqi Zhang$^{1}$, Xuanang Chen$^{2}$, \\
{\bfseries Ben He$^{2,3}$, Xin Xin$^{1}$, \bfseries Zhumin Chen$^{1}$, \bfseries Ying Zhou$^{1}$\thanks{Corresponding author.}} \\
\textsuperscript{1}Shandong University\\
\textsuperscript{2}Institute of Software, Chinese Academy of Sciences\\
\textsuperscript{3}University of Chinese Academy of Sciences\\
\texttt{fangan.dong@mail.sdu.edu.cn}, \texttt{yingzhou@sdu.edu.cn} 
}

\begin{document}
\maketitle
\begin{abstract}
Despite the strong reasoning capabilities of recent large language models (LLMs), achieving reliable performance on challenging tasks often requires post-training or computationally expensive sampling strategies, limiting their practical efficiency.
In this work, we first show that a small subset of neurons in LLMs exhibits strong predictive correlations with reasoning correctness.
Based on this observation, we propose \textbf{AdaRAS} (\textbf{Ada}ptive \textbf{R}easoning \textbf{A}ctivation \textbf{S}teering), a lightweight test-time framework that improves reasoning reliability by selectively intervening on neuron activations.
AdaRAS identifies Reasoning-Critical Neurons (RCNs) via a polarity-aware mean-difference criterion and adaptively steers their activations during inference, enhancing incorrect reasoning traces while avoiding degradation on already-correct cases.
Experiments on 10 mathematics and coding benchmarks demonstrate consistent improvements, including over 13\% gains on AIME-24 and AIME-25.
Moreover, AdaRAS exhibits strong transferability across datasets and scalability to stronger models, 
outperforming post-training methods with only minimal additional computation cost\footnote{The code and data are available at \url{https://github.com/cat-sk/AdaRAS}}.
\end{abstract}

\section{Introduction}

Recent large language models (LLMs)~\citep{DBLP:journals/corr/abs-2412-16720,DBLP:journals/corr/abs-2505-09388,DBLP:journals/corr/abs-2501-12948}
have demonstrated strong capability across a wide range of natural language processing tasks.
In particular, test-time scaling~\citep{DBLP:conf/nips/Wei0SBIXCLZ22,DBLP:journals/corr/abs-2407-21787} has improved LLM performance by allocating additional computation at inference, enabling applications in mathematical problem solving~\citep{DBLP:conf/icml/GuanZLSSZ0025,DBLP:journals/corr/abs-2501-19393}, code generation~\citep{DBLP:journals/corr/abs-2404-10952,DBLP:conf/iclr/JainHGLYZWSSS25}, and spatial reasoning~\citep{DBLP:journals/corr/abs-2403-11401}.
Despite these advances, reliable reasoning remains challenging even for state-of-the-art models, and most improvements rely on post-training methods~\citep{DBLP:journals/corr/abs-2402-03300,DBLP:journals/corr/abs-2502-01456,DBLP:journals/corr/abs-2503-14476} or costly test-time strategies such as prompt engineering~\citep{DBLP:conf/nips/TianPSJ0HMY24,DBLP:conf/icml/Saha0GWW25}, self-consistency~\citep{DBLP:conf/iclr/0002WSLCNCZ23,DBLP:journals/corr/abs-2410-16033}, or multi-step calibration~\citep{DBLP:journals/corr/abs-2110-14168,DBLP:journals/corr/abs-2503-04625,DBLP:journals/corr/abs-2408-03314}.
While effective, these approaches treat LLMs as black boxes, incur substantial inference overhead, and offer limited insight into the sources of reasoning errors.

\begin{figure}[t]
\centering
\includegraphics[width=\linewidth]{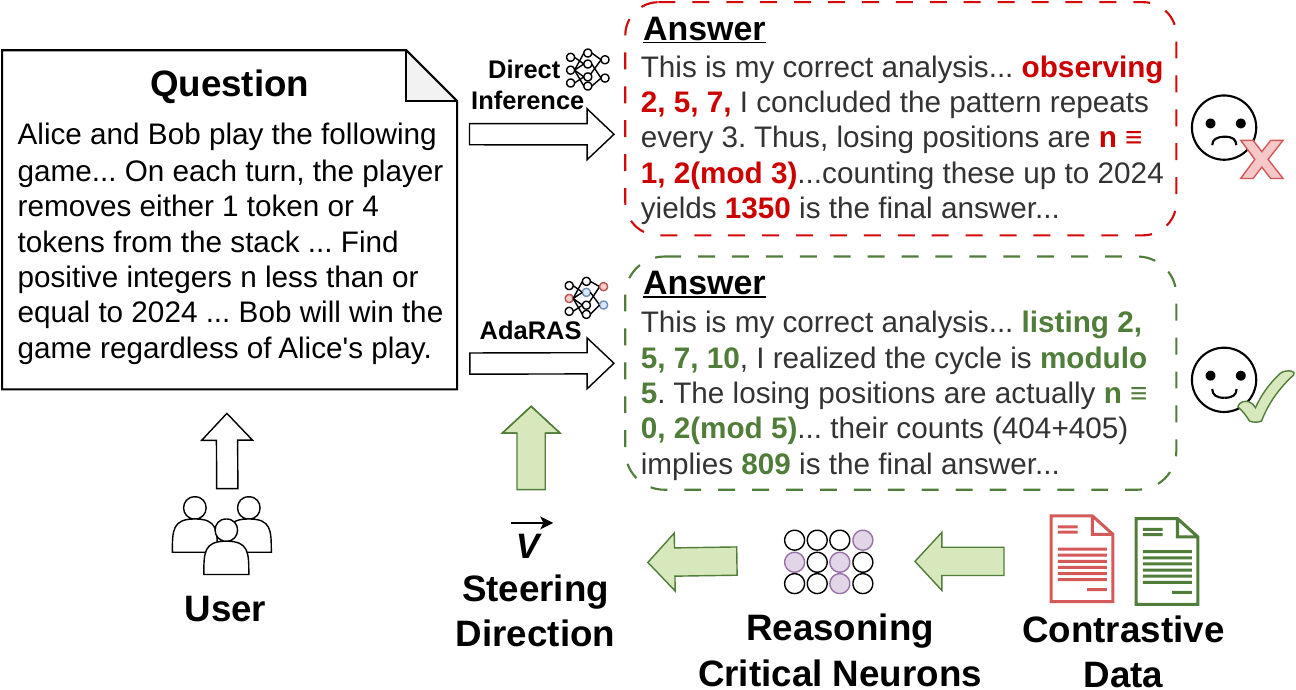}
\caption{An example of AdaRAS activation steering correcting an erroneous reasoning trajectory.
} 
\label{fig:overview}
\end{figure}

Recent research works in activation engineering~\citep{DBLP:journals/corr/abs-2308-10248} provides an internal alternative, showing that selectively manipulating activations in attention heads~\citep{DBLP:conf/nips/0002PVPW23} or MLP neurons~\citep{DBLP:conf/acl/RimskyGSTHT24} can control model behaviors, such as truthfulness~\citep{DBLP:journals/corr/abs-2310-06824}, refusal~\citep{DBLP:conf/iclr/LeePRMDND25}, or reducing bias~\citep{DBLP:journals/corr/abs-2402-00402}, without modifying model parameters or increasing inference cost.
This suggests that internal activations encode functionally specialized signals that can be directly leveraged for controllable behavior.
However, reasoning reliability is a fundamentally different problem: it is a trajectory-level property requiring temporally coherent, multi-step inference with long-range dependencies, rather than adjustment of a single output attribute.
It therefore remains unclear whether existing activation engineering techniques can identify and enhance the internal mechanisms underlying reasoning correctness.
This raises a central question: \textit{Can activation-level interventions improve LLM reasoning reliability?}

To address this question, we first examine whether reasoning correctness can be inferred via internal activations.
Inspired by probing studies of LLMs~\citep{DBLP:journals/tmlr/GurneeNPHTB23}, a pilot analysis reveals that a small subset of MLP neurons exhibits polarized activations between correct and incorrect reasoning trajectories, and that these activations are predictive of reasoning outcomes.
Building by this observation, we propose AdaRAS, an adaptive activation steering framework for test-time reasoning enhancement.
AdaRAS identifies RCNs by contrasting activations from correct and incorrect samples and applies polarity-based filtering to obtain a sparse, functionally consistent neuron set.
During inference, AdaRAS selectively intervenes on these RCNs only when a trajectory is likely incorrect, enabling targeted correction without degrading already-correct one.
Experiments on ten mathematics and coding benchmarks show that AdaRAS consistently improves accuracy across models, generalizes across tasks and datasets, and requires no additional training or sampling.

The contributions of this paper are threefold:
1) To our knowledge, we present the first systematic evidence that reasoning correctness can be predicted and improved through neuron interventions, establishing activation steering as a viable tool for enhancing LLM reasoning.
2) We introduce AdaRAS, a parameter-free, test-time activation steering framework that consistently enhances reasoning performance and transfers across tasks and datasets.
3) We provide mechanistic insights showing that AdaRAS stabilizes latent reasoning trajectories while preserving semantic representations, enabling plug-and-play deployment.
\section{Preliminaries}
This section presents preliminary results that specific neuron activations predict reasoning correctness, thereby motivating our proposed AdaRAS.

\subsection{Task Formulation}
\label{sec:task_formulation}
We first define \emph{Reasoning-Critical Neurons (RCNs)} as neurons who positively contribute to correct reasoning outcomes.
Given a dataset $D=\{(x_i, y_i)\}$ and a reasoning language model $\mathcal{M}$, we denote by $r_i$ the model-generated reasoning trace for input $x_i$, and by $a_i$ the final answer extracted from $r_i$. We define a binary correctness label $c_i = \mathbb{I}[a_i = y_i]$, which serves as the target signal throughout this work.
Architecturally, the model $\mathcal{M}$ consists of $L$ Transformer decoder layers. 
Following prior work~\citep{DBLP:conf/emnlp/GevaSBL21,DBLP:conf/nips/MengBAB22}, we focus on the MLP blocks, which are widely believed to encode high-level semantic patterns. 
Specifically, we take the intermediate activation for each MLP block as the internal representation.

\begin{figure}[t]
\centering
\includegraphics[width=0.9\columnwidth]{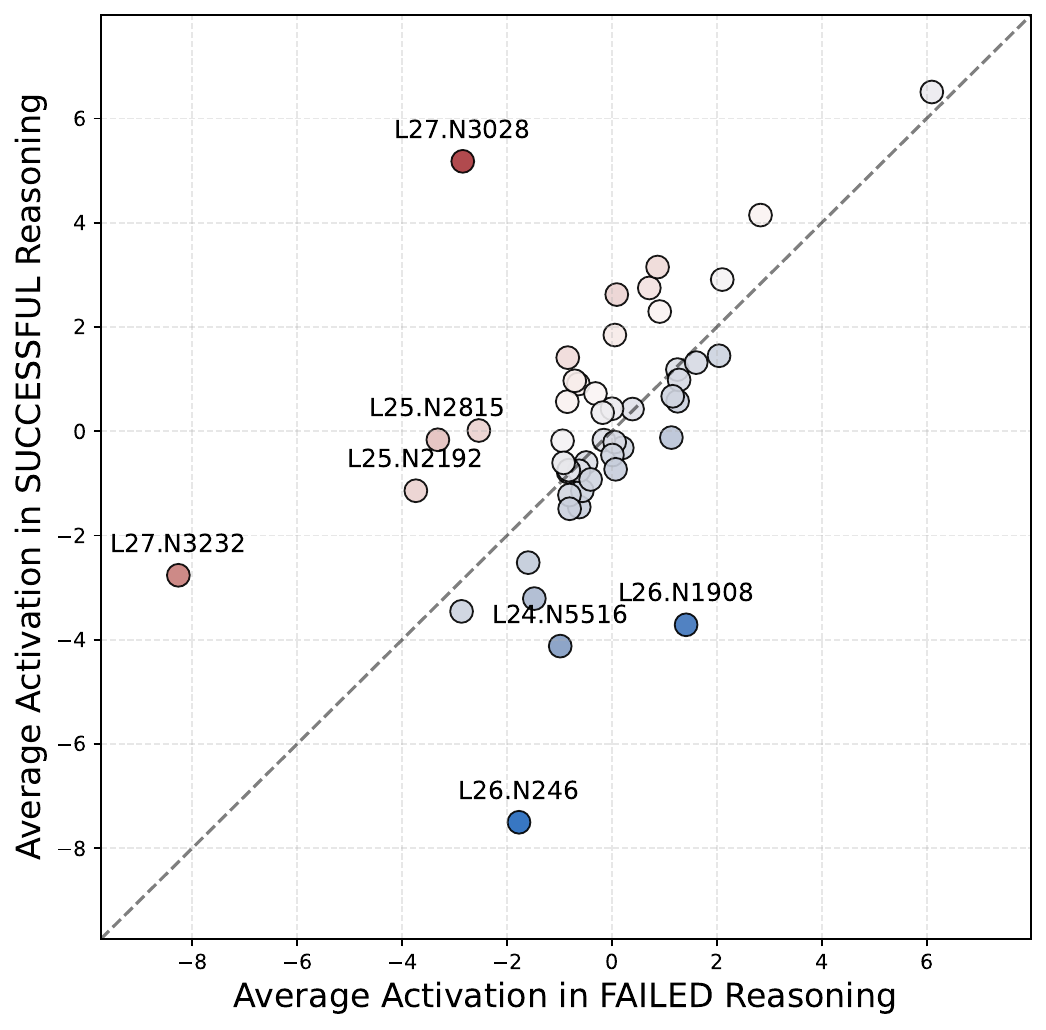} 
\caption{Comparison of activations of key neurons under successful and failed reasoning on AIME.}
\label{fig:pre:neuron_comparison}
\end{figure}

\subsection{LLM Reasoning Signatures}
\label{sec:pre:probing}
Motivated by recent advances in LLM interpretability~\citep{DBLP:journals/coling/Belinkov22,DBLP:journals/tmlr/GurneeNPHTB23,DBLP:conf/coling/Ju0DYRL24}, we ask whether reasoning correctness can be inferred directly from intermediate neuron activations at test time.
As a preliminary study, we probe last-token activations of Qwen3 series models~\citep{DBLP:journals/corr/abs-2505-09388} on mathematic datasets (i.e., AIME and AMC-12) to assess their predictive power for reasoning reliability.
In contrast to prior work on stylistic control or knowledge editing, we focus on identifying neurons that are specifically predictive of reasoning correctness.
Detailed experimental setups are provided in the Appendix~\ref{app:preliminary}.

\begin{table}[t!]
\centering
\caption{AUROC of probing classifiers trained on last-token activations for predicting reasoning correctness.}
\label{tab:pre:probing}
\begin{tabular}{llc}
\toprule
\textbf{Dataset} & \textbf{Model} & \textbf{AUROC} \\
\midrule
\multirow{2}{*}{AIME (24+25)} & Qwen3-1.7B & 0.7639 \\
& Qwen3-4B & 0.7153 \\
\midrule
\multirow{2}{*}{AMC-12} & Qwen3-1.7B & 0.7091 \\
& Qwen3-4B & 0.6727 \\
\bottomrule
\end{tabular}%
\end{table}
\begin{figure*}[t]
\centering
\includegraphics[width=\textwidth]{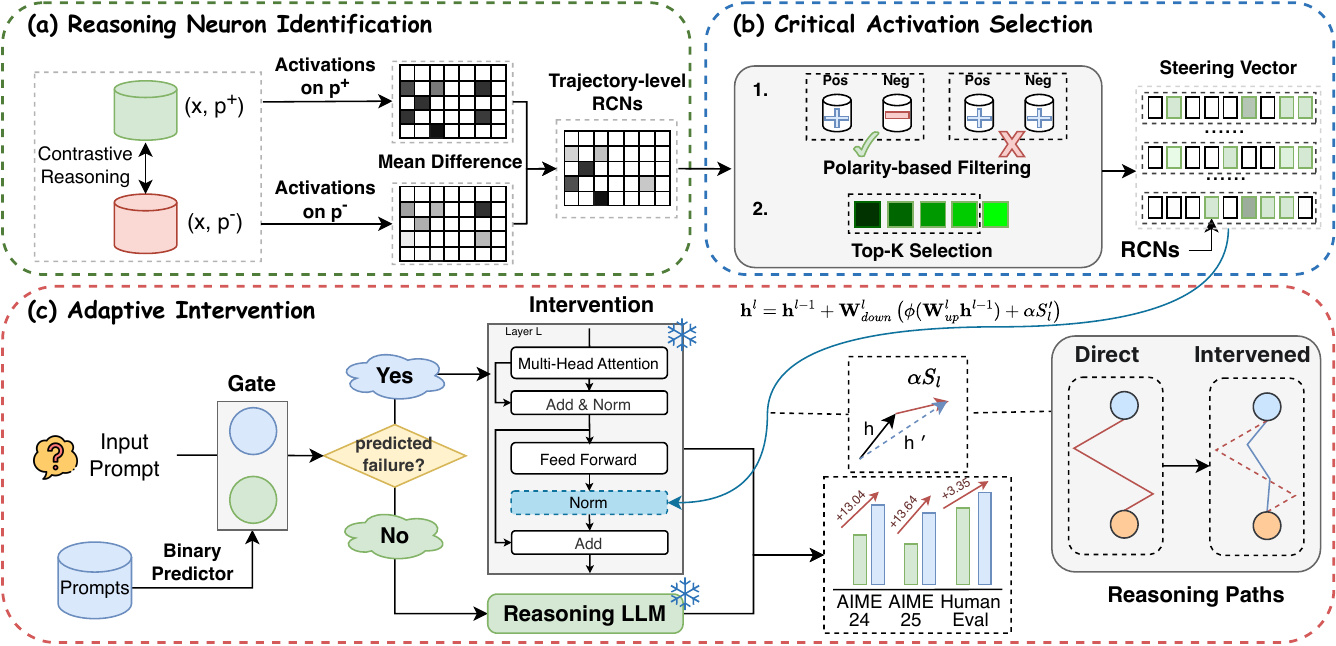}
\caption{Overview of AdaRAS:
(1) reasoning neuron identification (\secref{sec:method:identify}),
which identifies critical neurons by measuring global activation differences between contrastive reasoning trajectories;
(2) critical activation selection (\secref{sec:method:steer}), 
which further refines RCNs based on activation polarity variations;
(3) adaptive intervention (\secref{sec:method:adaptive}), 
which enhances the reliability of steering by predicting reasoning failures and performing adaptive interventions.
}
\label{fig:framework}
\end{figure*}

\paragraph{Finding 1. LLM reasoning traces leading to correct versus incorrect answers exhibit distinct activation patterns.}
As shown in Figure~\ref{fig:pre:neuron_comparison}, we compare the mean activations of highly contrastive neurons under successful and failed reasoning.
Notably, the 3028-th neuron in layer 27 shows substantially higher activation during successful reasoning, whereas the 246-th and 1908-th neurons in layer 26 exhibit the opposite trend.
This indicates that specific neurons capture informative signals associated with reasoning success at test time.

\paragraph{Finding 2. Token-level neuron activations are predictive of the final correctness of LLM reasoning.}
As reported in Table~\ref{tab:pre:probing}, probing classifiers trained solely on last-token activations achieve AUROC scores around 0.7 across two datasets and two models, reaching up to 0.76 on AIME with the Qwen3-1.7B model.
These observations suggest that neuron activation states are systematically correlated with reasoning outcomes.

\section{Methodology}

Figure~\ref{fig:framework} overviews our AdaRAS framework. 
First, we identify RCNs by contrasting activation patterns from correct and incorrect reasoning samples (\secref{sec:method:identify}).
Then, we apply sparse activation steering to selectively intervene on these neurons (\secref{sec:method:steer}).
To avoid perturbing already-correct reasoning, we introduce an adaptive intervention strategy that conditionally modulates activations at test time (\secref{sec:method:adaptive}).
Finally, we describe the construction of parallel reasoning traces used in our experiments (\secref{sec:method:parallel}).

\subsection{Reasoning Neuron Identification}
\label{sec:method:identify}
We identify reasoning-critical neurons (RCNs) by directly contrasting activation patterns between correct and incorrect reasoning trajectories.
While probing-based methods can assign neuron importance via classifier weights, we find them unreliable in practice due to their sensitivity to feature selection and dependence on probe generalization.
Following prior works~\citep{DBLP:conf/acl/RimskyGSTHT24}, we therefore adopt a probing-free, activation-level criterion based on mean activation differences.

Let $a_i^l(t)$ denote the activation value of neuron $i$ at layer $l$ for token $t$.
Given paired correct and incorrect reasoning traces $\mathcal{P} = \{(p_k^+, p_k^-)\}_{k=1}^N$, we define the reasoning-critical score of neuron $(l,i)$ as the difference in mean activations:
\begin{equation}
\begin{gathered}
S(l,i)
=
\EV_{k \sim [N]}
\Big[
\mu(a_i^l, p_k^+) - \mu(a_i^l, p_k^-)
\Big], \\
\mu(a, p)
\triangleq
\frac{1}{|p|}
\sum_{t \in p} a(t),
\end{gathered}
\end{equation}
where $|p|$ denotes the length of the reasoning trace.
Unlike prior works~\citep{DBLP:conf/acl/RimskyGSTHT24} which adopt token-level MD applied to the final answer, this global formulation captures sustained activation patterns accumulated throughout the reasoning process.
Specifically, for each layer $l$, the scores of all neurons form a layer-wise vector $\mathbf{S}_l = [S(l,1), \dots, S(l,d)]^\top$,
which serves as the steering direction for test-time intervention.

\subsection{Critical Activation Selection}
\label{sec:method:steer}
From preliminary results, our key observation is that correct reasoning is supported by structured activation patterns formed by a small subset of neurons, rather than uniformly distributed across entire layers.
Therefore, instead of intervening at the layer level, we perform neuron-level selection and steering.
Specifically, most modern LLMs, such as LLaMA and Qwen, employ SwiGLU activations~\citep{DBLP:journals/corr/abs-2002-05202}, whose signed output is necessary for representations learning~\citep{DBLP:journals/corr/abs-2405-20768}.
Our empirical analysis also reveals that neurons whose average activation polarity differs between correct and incorrect reasoning trajectories are particularly discriminative (see \ref{sec:exp:ablation} for detailed results).
Intuitively, such polarity reversals indicate neurons that actively support or suppress valid reasoning, depending on the reasoning outcome.

Formally, for each neuron $(l,i)$, we retain it for steering only if
\begin{equation}
\EV_k\!\left[\mu(a_i^l, p_k^+)\right]
\cdot
\EV_k\!\left[\mu(a_i^l, p_k^-)\right]
< 0,
\end{equation}
i.e., its mean activation exhibits a sign flip between correct and incorrect traces.
This polarity-based filtering yields a sparse candidate set of RCNs.
We then rank the retained neurons by the magnitude of their importance scores $|S(l,i)|$ and select the top-$K$ neurons to construct a sparse steering vector $\bm S'_l$ for each layer.
The steering strength is controlled by a positive scalar $\alpha \in [0,\infty]$.
Activation steering is applied within the MLP block of the LLM:
\begin{equation}
\mathbf{h}^{l}
=
\mathbf{h}^{l-1}
+
\mathbf{W}_{down}^{l}
\left(
\phi(\mathbf{W}_{up}^{l} \mathbf{h}^{l-1})
+
\alpha \bm S'_l
\right),
\label{eq:mlp_intervention}
\end{equation}
where $\bm S'_l$ is constructed once from a reference dataset and applied uniformly at each decoding step during test-time inference.

\subsection{Adaptive Intervention Strategy}
\label{sec:method:adaptive}
Our early experiments shown a trade-off in activation steering: while it effectively rectifies incorrect reasoning, it inadvertently degrade performance on some originally correct samples.
To address this, we introduce a lightweight failure prediction module that adaptively gates steering at test time.
Specifically, we formulate a prediction task to estimate whether the base model can correctly answer a given input.
The predictor operates on early neuron activations induced by the input prompt, which capture the model's initial reasoning state.
To reduce dimensionality, we select a small set of activations using $F$-statistics method as the input features.
A attention-based classifier is then trained on these features and achieves an AUROC of 0.8347 on AIME dataset, which further supports our insights that there exists a strong correlation between activation patterns and reasoning outcomes.
During inference, the predictor serves as a gate: AdaRAS is applied only when a reasoning failure is predicted.
This adaptive intervention improves overall reasoning reliability while preserving performance on inputs that are already correctly handled.

\subsection{Contrastive Data Construction}
\label{sec:method:parallel}
As mentioned in Section~\ref{sec:method:identify}, identifying RCNs requires contrastive reasoning traces that share the same input but differ in outcome correctness.
To this end, we construct paired positive and negative reasoning trajectories via self-sampling.
For each input prompt, we sample multiple reasoning trajectories using high-temperature decoding and partition them into positive and negative sets based on answer correctness.
Contrastive pairs are then formed by matching traces with identical inputs but opposite outcomes.
Aggregating all pairs yields a contrastive reasoning set $\mathcal{P} = \{(p^+, p^-)\}$, which is used to identify and transfer RCN activations.
\begin{table*}[t]
\centering
\caption{Experimental results on mathematics and coding benchmarks.
We compare AdaRAS with probing-based steering and post-training baselines, evaluated using the Accuracy metric.
\textbf{Bold} numbers denote the best performance on each dataset.
\cb{Red subscripts} indicate the improvement of AdaRAS over the CoT prompting.}
\label{tab:exp:main}
\resizebox{\textwidth}{!}{%
\begin{tabular}{ll c ccc cc}
\toprule
\multirow{2}{*}{\textbf{Task}} & \multirow{2}{*}{\textbf{Dataset}} & & \multicolumn{3}{c}{\textbf{Post-training}} & \multicolumn{2}{c}{\textbf{Steering}} \\
\cmidrule(lr){4-6} \cmidrule(lr){7-8}
& & CoT & R1-Distill & Nemotron & OpenThinker & Probing & AdaRAS \\
\midrule
\multirow{6}{*}{Math} 
& AIME-24 & 47.83 & 34.78 & 39.13 & 26.08 & 43.48 & \textbf{60.87}$_{\cb{+13.04}}$ \\
& AIME-25 & 40.91 & 22.73 & 40.91 & 50.00 & 50.00 & \textbf{54.55}$_{\cb{+13.64}}$ \\
& AIME-Extend & 47.33 & 21.33 & 47.33 & 42.67 & 48.00 & \textbf{52.67}$_{\cb{+5.34}}$ \\
& MATH-500 & 84.80 & 66.20 & 85.40 & 82.00 & 85.40 & \textbf{86.40}$_{\cb{+1.60}}$ \\
& GSM8K & 88.32 & 72.93 & 87.57 & 67.55 & 88.32 & \textbf{89.08}$_{\cb{+0.76}}$ \\
& AMC-12 & 65.93 & 41.76 & 65.93 & 53.85 & \textbf{73.63} & 70.33$_{\cb{+4.40}}$ \\
\midrule
\multirow{4}{*}{Code} 
& HumanEval & 77.18 & 45.64 & 57.72 & 59.06 & 77.85 & \textbf{79.19}$_{\cb{+2.01}}$ \\
& HumanEval+ & 69.80 & 42.95 & 51.68 & 53.69 & 72.48 & \textbf{73.15}$_{\cb{+3.35}}$ \\
& MBPP & 68.78 & 42.59 & 53.97 & 35.71 & 70.37 & \textbf{72.22}$_{\cb{+3.44}}$ \\
& MBPP+ & 58.20 & 37.30 & 43.92 & 30.95 & 59.52 & \textbf{60.58}$_{\cb{+2.38}}$ \\
\bottomrule
\end{tabular}%
}
\end{table*}

\section{Experiments}

\subsection{Experimental Setup}


\paragraph{Dataset.}
We evaluate AdaRAS on 10 widely used mathematics and coding benchmarks that require coherent reasoning.
These include standard competition-level benchmarks such as AIME-24/25 and AMC-12, as well as large-scale datasets such as GSM8K~\citep{DBLP:journals/corr/abs-2110-14168} and MBPP+~\citep{evalplus}.
For datasets lacking an official training split, we partition the data into a small probing set and a remaining set reserved for test-time steering evaluation. 
For each probing sample, we generate 4 pairs of positive and negative reasoning traces to identify RCNs. 
Dataset statistics are summarized in Table~\ref{tab:dataset_stats}.

\paragraph{Baseline.}
We compare AdaRAS with the following baselines:
(1) \textbf{Chain-of-Thought (CoT)}. We include the vanilla CoT prompting performance of Qwen3-1.7B as a baseline, on which AdaRAS is applied.
(2) \textbf{Post-trained reasoning models}. Post-training methods (e.g., SFT, PPO, GRPO) are known to enhance model's reasoning performance. We therefore consider three publicly available post-trained models with comparable scales: DeepSeek-R1-Distill-Qwen-1.5B~\citep{DBLP:journals/corr/abs-2501-12948}, OpenThinker3-1.5B~\citep{DBLP:journals/corr/abs-2506-04178}, and OpenReasoning-Nemotron-1.5B~\citep{DBLP:journals/corr/abs-2504-01943}.
(3) \textbf{Probing-based steering}. We use the weights of the probing classifier trained in \secref{sec:pre:probing} as neuron importance scores, select the same number of top-ranked neurons as in AdaRAS, and perform activation addition steering~\citep{DBLP:journals/corr/abs-2308-10248} accordingly. This baseline is denoted as \textit{Probing}.

\paragraph{Evaluation.}
All methods are evaluated under the same greedy decoding setup for fair comparison; see Appendix~\ref{app:imp:prompts} for details.
Following prior work, we report Accuracy as the metric across all datasets.

\subsection{Main Results}
\label{sec:exp:main}
Table~\ref{tab:exp:main} compares AdaRAS against the CoT and post-training methods across ten benchmarks.

\paragraph{AdaRAS consistently improves reasoning correctness across all datasets, even surpassing post-trained LLMs.}
On average, AdaRAS achieves an improvement of $\sim$5\% over CoT-based inference.
Notably, on the challenging AIME-25 benchmark, AdaRAS yields a substantial gain of 13.64\%, surpassing all existing open-source post-trained models of comparable scale.
We further observe that the performance gains introduced by AdaRAS are more pronounced on hard reasoning benchmarks, such as AIME and AMC. 
On these datasets, AdaRAS achieves an average improvement of 9.11\%, compared to a more modest gain of 2.26\% on relatively easier benchmarks.
These results suggest that steering RCN activations toward favorable states can substantially enhance the reliability of the generation process, particularly for complex and cognitively demanding reasoning tasks.

\paragraph{Probing-based activation steering is inherently unstable.}
As shown in Table~\ref{tab:exp:main}, vanilla probing-based steering yields mixed results across datasets.
While it achieves notable gains on AMC-12, even surpassing AdaRAS, it fails to generalize and can be detrimental on more challenging benchmarks.
In particular, on AIME-24, probing-based steering degrades accuracy by 4.35\% compared to the unsteered CoT baseline.
These results indicate that naive probing-based approaches are highly sensitive to neuron selection, and may harm reasoning performance without adaptive intervention mechanisms, which are explicitly addressed in AdaRAS.

\begin{table}[t]
\centering
\caption{Transferability results of AdaRAS across datasets.
We compare in-domain steering using RCNs identified on each dataset (\textit{ID. RCNs}) with cross-dataset steering using RCNs identified on AIME (\textit{AIME-RCNs}).
}
\label{tab:exp:transferability}
\begin{tabular}{lcc}
\toprule
\textbf{Dataset} & \textbf{ID. RCNs} & \textbf{AIME-RCNs} \\
\midrule
MATH-500 & 86.40 & \textbf{87.40} \\
GSM8K   & 89.08 & \textbf{89.39} \\
AMC-12   & 70.33 & \textbf{71.43} \\
\midrule
HumanEval   & 79.19 & \textbf{80.54} \\
HumanEval+ & 73.15 & 73.15 \\
MBPP        & 72.22 & \textbf{72.75} \\
MBPP+      & 60.58 & \textbf{61.11} \\
\bottomrule
\end{tabular}%
\end{table}

\subsection{Generalizability}
Tables~\ref{tab:exp:transferability} and~\ref{tab:exp:scalability} show the transferability and generalization of AdaRAS, covering cross-dataset setting and larger model scale.

\paragraph{RCNs exhibit strong cross-dataset and cross-task transferability.}
As shown in Table~\ref{tab:exp:transferability}, compared to using dataset-specific RCNs, we observe that cross-dataset activation steering with RCNs identified on AIME consistently yields better performance, resulting in small but stable gains, within 1\%, across all evaluated datasets.
Notably, these RCNs also transfer across task domains: when applied to coding benchmarks, they continue to improve performance.
These results support our motivation for identifying RCNs and suggest that such neurons capture task-agnostic reasoning mechanisms.
In particular, RCNs identified from more challenging tasks appear to generalize broadly to diverse reasoning settings.

\begin{table}[t]
\centering
\caption{Scalability results of AdaRAS on Qwen3-4B.
We evaluate AdaRAS on a larger base model using cross-dataset steering with RCNs identified on AIME.}
\label{tab:exp:scalability}
\begin{tabular}{lcc}
\toprule
\textbf{Dataset} & \textbf{Qwen3-4B} & \textbf{+AdaRAS} \\
\midrule
AIME-24 & 56.52 & \textbf{60.87} \\
AIME-25 & 59.09 & \textbf{72.73} \\
AIME-Extend & 68.67 & \textbf{76.67} \\
MATH-500 & 90.00 & \textbf{91.20} \\
GSM8K & 92.72 & \textbf{93.93} \\
AMC-12 & 76.92 & \textbf{80.22} \\
\midrule
HumanEval & 91.28 & \textbf{92.62} \\
HumanEval+ & 81.88 & \textbf{84.56} \\
MBPP & 79.37 & \textbf{82.28} \\
MBPP+ & 68.78 & \textbf{69.31} \\
\bottomrule
\end{tabular}%
\end{table}

\paragraph{AdaRAS scales effectively to stronger reasoning models.}
Table~\ref{tab:exp:scalability} shows that AdaRAS continues to yield performance improvements when applied to the larger and more capable reasoning model.
Even when the base model already achieves high accuracy, whereas Qwen3-4B surpasses 90\% accuracy on relatively easier benchmarks such as MATH-500, GSM8K, and HumanEval, AdaRAS still delivers consistent accuracy gains of around 1\%.
Notably, on more challenging datasets such as AIME, the improvements are more pronounced, mirroring the trends observed in previous experiments.
Overall, these results indicate that AdaRAS acts as a complementary test-time intervention, providing consistent gains even as the base model's reasoning capability improves.

Moreover, additional results in the Appendix~\ref{app:additional} show that AdaRAS remains effective against stronger test-time baselines, scales to larger LLMs, exhibits stability across prompt variants, and incurs minimal inference and offline training overhead.

\subsection{Ablation Studies}
\label{sec:exp:ablation}
Table~\ref{tab:exp:ablation} reports the ablation results for AdaRAS.
(1) Random Steering consistently degrades performance, indicating that indiscriminate neuron intervention is harmful.
(2) Removing the MD-based estimation results in the sharpest decline on AIME-24 (i.e., 60.87\% $\to$ 43.48\%), dropping even below the unsteered baseline. This underscores the necessity of MD for accurately identifying RCNs.
(3) Disabling polarity-based activation selection causes an approximate 10\% accuracy drop on both AIME-24 and AIME-25, highlighting the importance of discriminative activation polarity.
(4) Omitting adaptive intervention degrades overall performance by interfering with originally correct reasoning trajectories.
Overall, all components are essential to the effectiveness of AdaRAS; 
limited by space, further analysis is in Appendix~\ref{app:additional:ablation}.

\begin{table}[t]
\centering
\caption{Ablation results of key components in AdaRAS.
\textit{Random Steering} applies steering to randomly sampled neurons.
\textit{AdaRAS w/o MD} reduces the method to probing-based neuron importance estimation, instead of Mean Difference designed in \secref{sec:method:identify}.
\textit{AS} and \textit{AI} denote Activation Selection and Adaptive Intervention, corresponding to the designs in \secref{sec:method:steer} and \secref{sec:method:adaptive}, respectively.
}
\label{tab:exp:ablation}
\begin{tabular}{lcc}
\toprule
\textbf{Method} & \textbf{AIME-24} & \textbf{AIME-25} \\
\midrule
Qwen3-1.7B & 47.83 & 40.91 \\
\midrule
Random Steering & 34.78 & 27.27 \\
\midrule
AdaRAS w/o MD & 43.48 & 50.00 \\
AdaRAS w/o AS & 52.17 & 45.45 \\
AdaRAS w/o AI & 56.52 & 45.45 \\
AdaRAS & \textbf{60.87} & \textbf{54.55} \\
\bottomrule
\end{tabular}%
\end{table}
\section{Analysis}

\begin{figure}[t]
\centering

\begin{subfigure}[b]{0.49\linewidth}
\centering
\includegraphics[width=\linewidth]{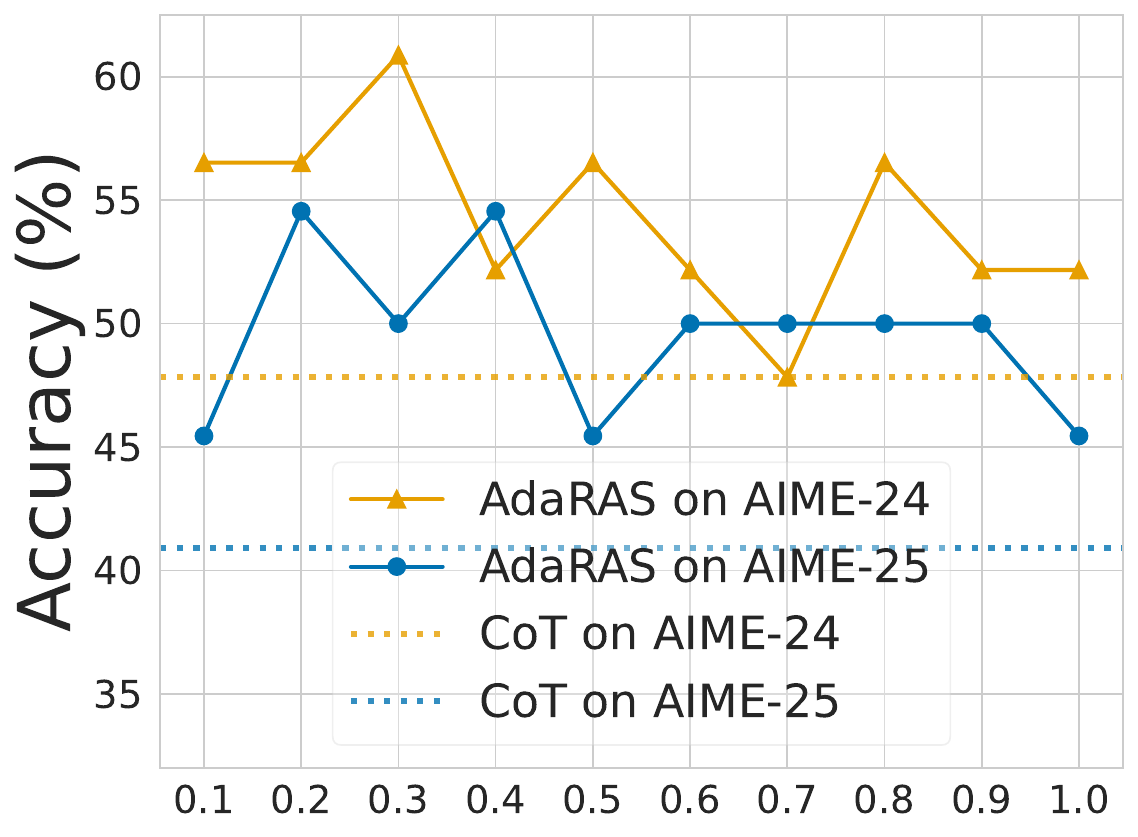}
\caption{Intervention strength $\alpha$}
\label{fig:analysis:alpha}
\end{subfigure}
\hfill
\begin{subfigure}[b]{0.49\linewidth}
\centering
\includegraphics[width=\linewidth]{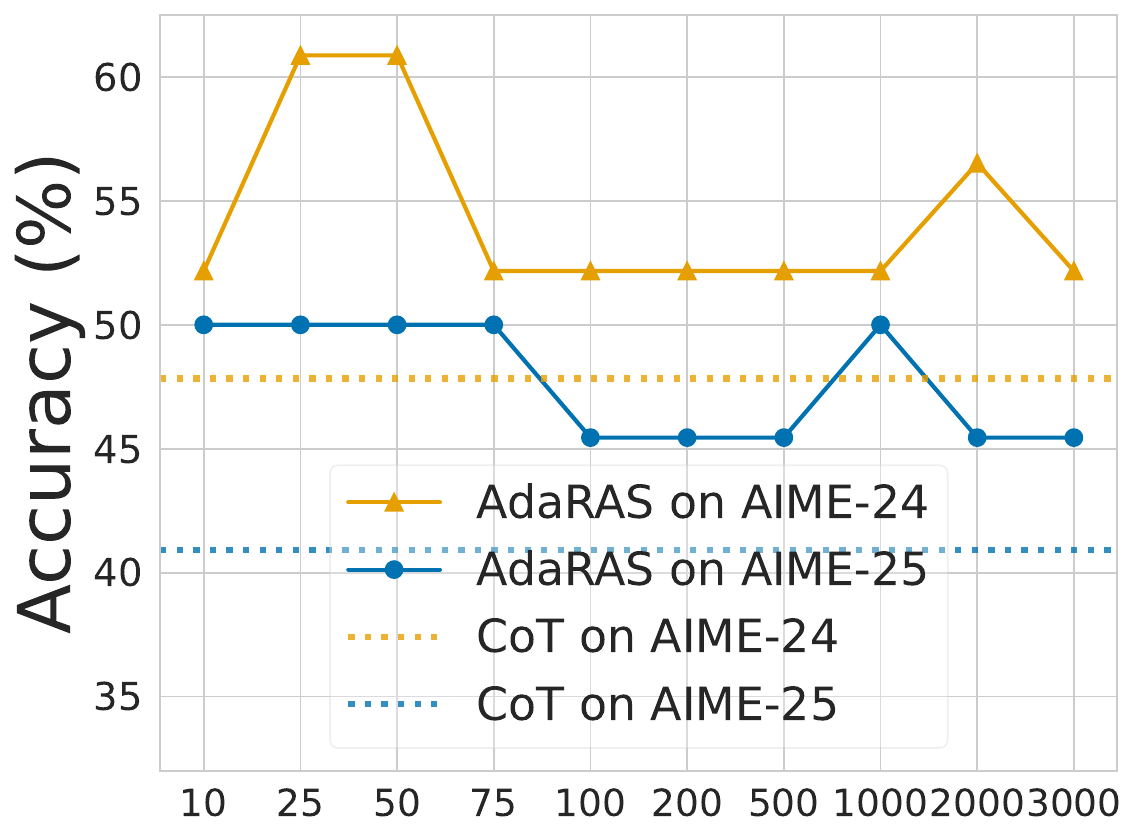}
\caption{Top-$K$ neurons}
\label{fig:analysis:top_k}
\end{subfigure}

\caption{Effect of hyperparameter.
}
\label{fig:analysis:hyperparams}
\end{figure}
\begin{figure}[t]
\centering
\begin{subfigure}[b]{0.48\columnwidth}
\centering
\includegraphics[width=\linewidth]{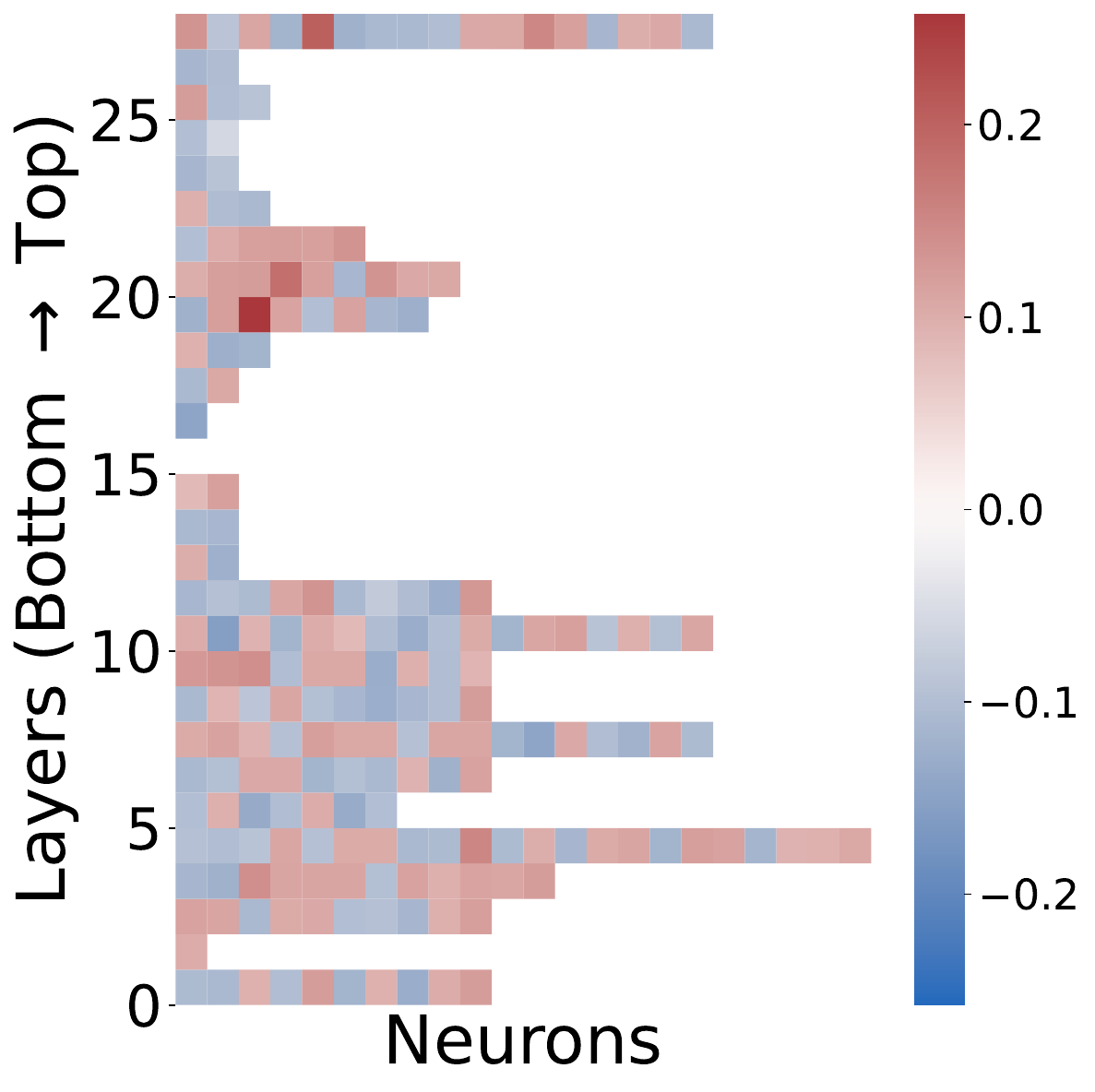}
\caption{Probing-based steering}
\label{fig:analysis:heatmap_probing}
\end{subfigure}
\hfill
\begin{subfigure}[b]{0.48\columnwidth}
\centering
\includegraphics[width=\linewidth]{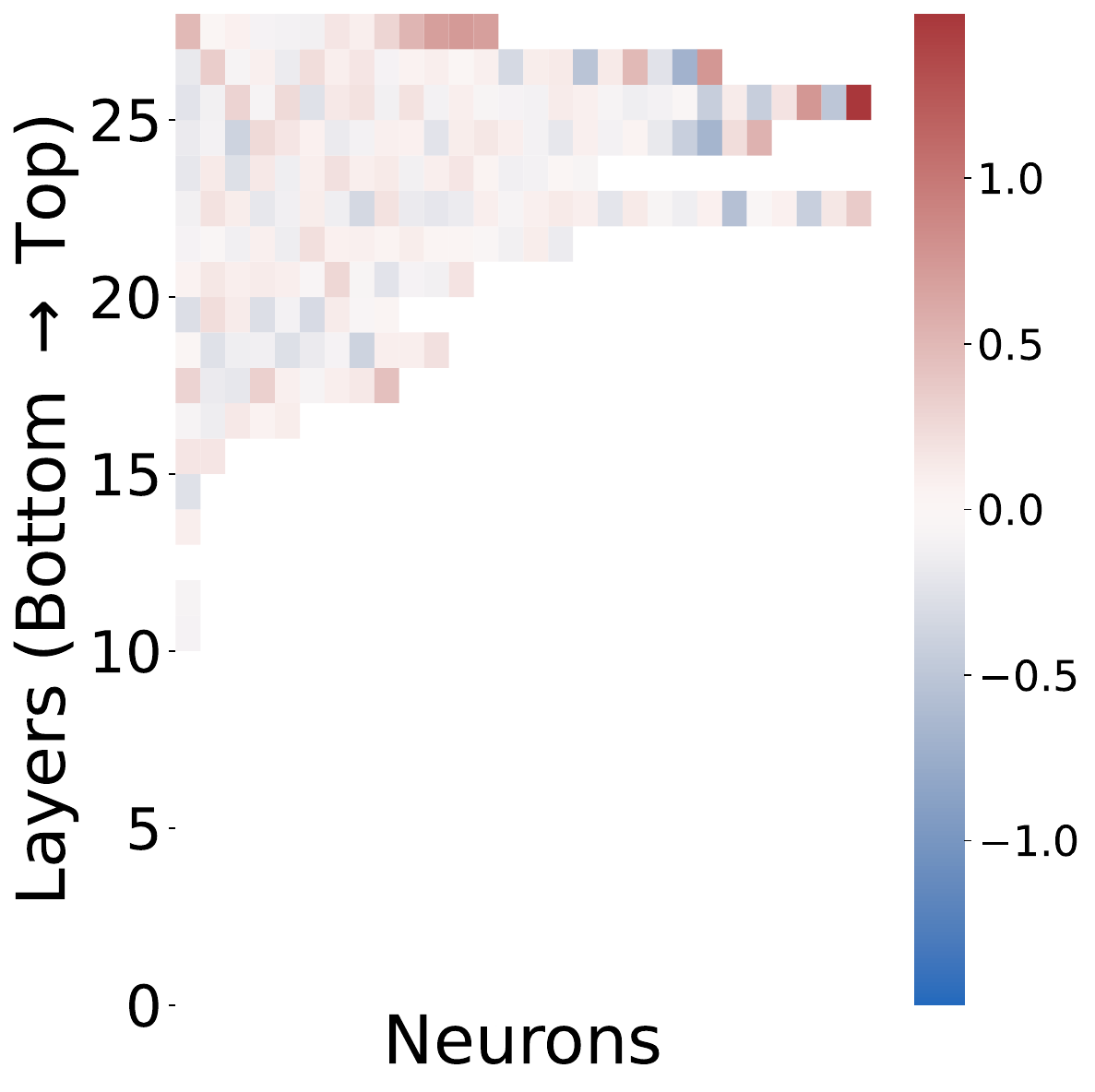}
\caption{AdaRAS}
\label{fig:analysis:heatmap_adaras}
\end{subfigure}
\caption{Visualization of activation shifts induced by steering. Neurons are sorted by their indices in layer, and unsteered neurons are omitted for clarity.}
\label{fig:analysis:heatmap}
\end{figure}

\subsection{Effect of Hyperparameter}
\label{sec:analysis:params}

In this section, we vary the intervention strength $\alpha \in [0,1]$ in Eq.\ref{eq:mlp_intervention} to control the activation shift magnitude. 
We further vary the number of selected neurons to assess the impact of RCN sparsity.

\paragraph{Effect of intervention strength.}
As shown in Figure~\ref{fig:analysis:alpha}, AdaRAS achieves peak performance at $\alpha = 0.3$ on AIME-24 and $\alpha = 0.4$ on AIME-25.
Increasing $\alpha$ beyond these values leads to performance decline, suggesting that excessive intervention interferes with the model's reasoning process.
Notably, performance drops to the unsteered baseline only when $\alpha = 0.7$ on AIME-24.
This suggests that AdaRAS is relatively robust to suboptimal hyperparameter choices and does not significantly harm the original reasoning capability under most settings.

\paragraph{Effect of top-$K$ RCNs selection.}
In the ablation study, we have shown that removing AdaRAS's polarity-aware filtering of RCNs leads to degraded steering performance.
Figure~\ref{fig:analysis:top_k} further reveals the sparse nature of RCNs.
Specifically, steering performance improves as more RCNs are intervened on, peaks at approximately top-50 neurons (about 0.03\% of all neurons), and degrades thereafter.
In particular, when intervening on top-2000 neurons (about 1.20\%), performance on AIME-24 drops by around 6\% compared to performance at top-50.
These results provide strong empirical evidence supporting our hypothesis in \secref{sec:method:steer} that RCNs are sparse and highly selective, and that indiscriminate intervention on a large set of neurons can be detrimental to reasoning performance.

\begin{figure*}[t]
\centering

\begin{subfigure}[b]{0.48\textwidth}
\centering
\includegraphics[width=\textwidth]{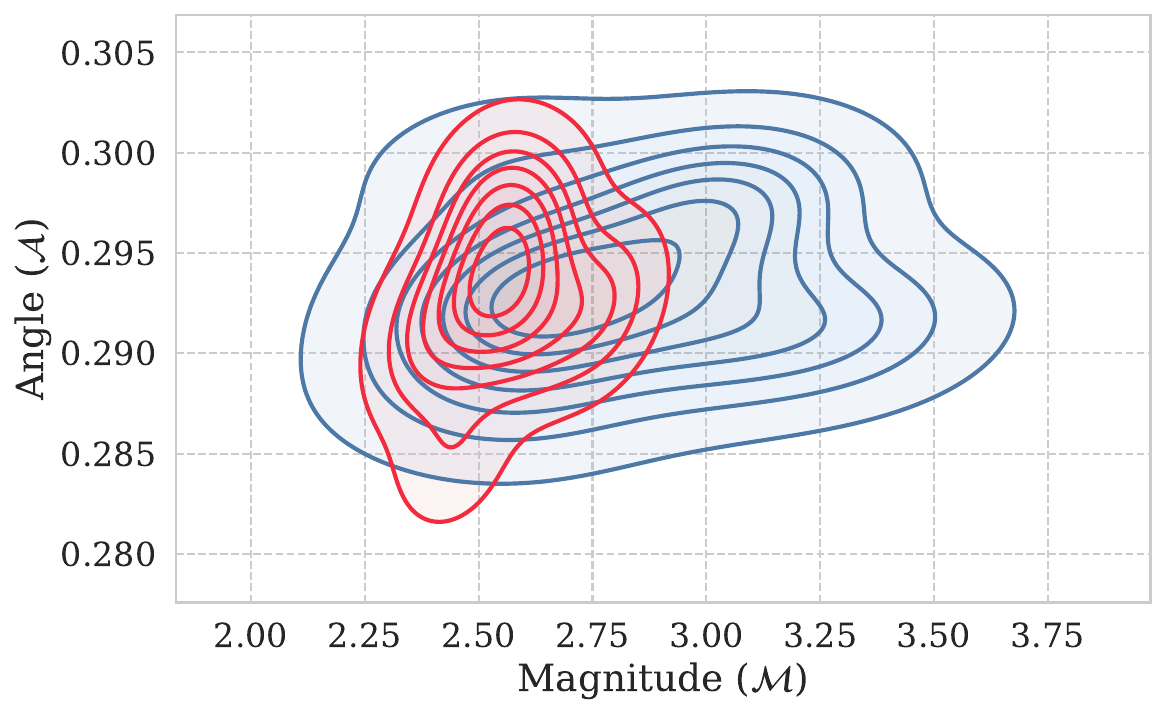}
\caption{AIME (In-domain steering)}
\label{fig:analysis:feats_aime}
\end{subfigure}
\hfill
\begin{subfigure}[b]{0.48\textwidth}
\centering
\includegraphics[width=\textwidth]{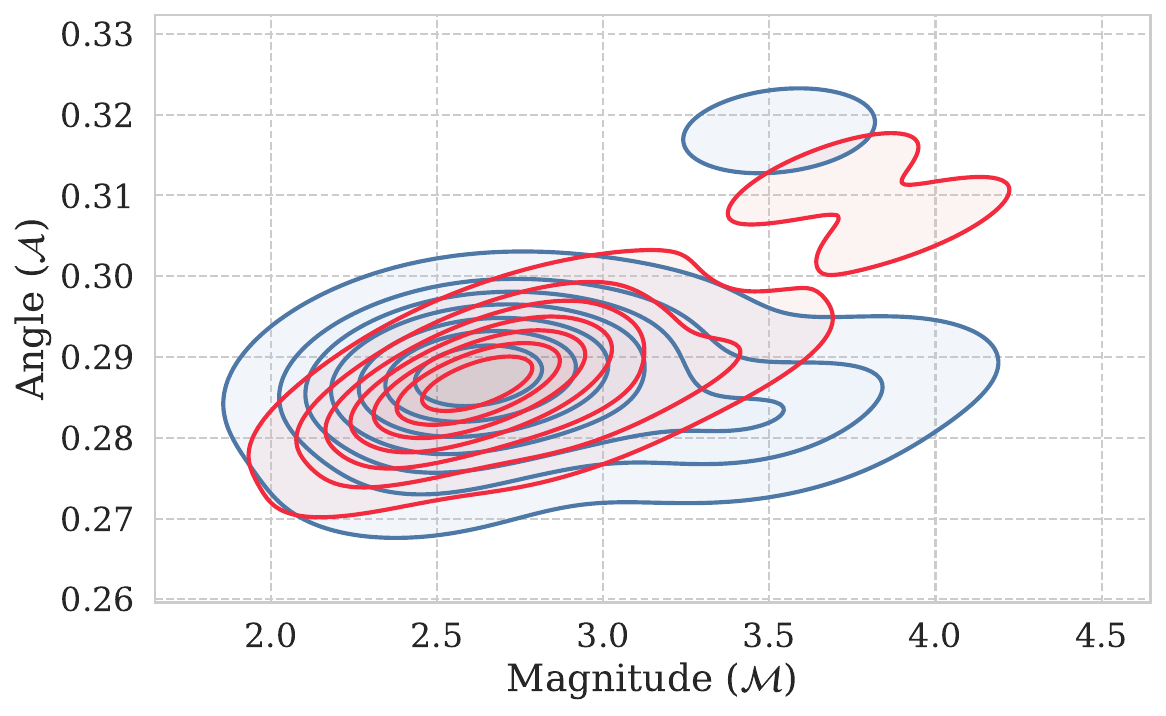}
\caption{MBPP (Cross-task steering)}
\label{fig:analysis:feats_mbpp}
\end{subfigure}

\caption{Feature distribution discrepancy of activations \textcolor[HTML]{4e79a7}{before} and \textcolor[HTML]{f22b3f}{after} intervention.
The x-axis denotes the magnitude feature, and the y-axis denotes the angle feature of activations.
Larger values indicate greater deviations in activation space.
Detailed definitions are provided in Appendix~\ref{app:reasoning_feat}.
}
\label{fig:analysis:feats}
\end{figure*}

\subsection{Visualization of Steering}
\label{sec:analysis:vis}

We visualize the effects of AdaRAS from neuron-level activation shifts to trajectory-level dynamics.

\paragraph{AdaRAS mainly intervenes on later layer neurons.}
Figure~\ref{fig:analysis:heatmap} compares activation shifts of RCNs induced by vanilla probing-based steering and AdaRAS on AIME-24 and AIME-25, where shifts are computed as the difference between mean neuron activations before and after steering.
Compared to probing-based methods, which introduce widespread changes in early layers, AdaRAS concentrates its interventions on later layers and induces substantially smaller activation shifts, with most values tightly centered around zero.
This observation aligns with prior findings~\citep{DBLP:conf/emnlp/BiranGYGG24} that later layer neurons are more important for multi-hop reasoning.

\paragraph{AdaRAS stabilizes latent reasoning trajectories without altering semantic modeling.} 
To further analyze the effect of AdaRAS on the reasoning process, we examine the stability of latent reasoning paths.
Following~\citet{DBLP:conf/iclr/00110YW025}, we compute two trajectory-level features from layer-wise activations: \textbf{magnitude} and \textbf{angle}, which capture reasoning path curvature and semantic modeling level, respectively.
Formal definitions are provided in the Appendix~\ref{app:reasoning_feat} due to space limitations.
As shown in Figure~\ref{fig:analysis:feats}, under both in-domain and cross-dataset steering, steered samples exhibit significantly lower and more concentrated magnitude values than unsteered samples, indicating that AdaRAS reduces fluctuations in latent reasoning trajectories.
In contrast, the angle metric remains largely unchanged, suggesting that AdaRAS stabilizes reasoning dynamics without substantially altering semantic modeling behavior.
\section{Related Works}

\paragraph{Reasoning Reliability.}

Improving reasoning reliability in LLMs has been widely studied via test-time scaling.
Prior work archives performance gains via post-training methods~\citep{DBLP:conf/nips/Ouyang0JAWMZASR22, DBLP:conf/iclr/LightmanKBEBLLS24,DBLP:journals/corr/abs-2402-03300} or costly test-time techniques, like prompt engineering~\citep{DBLP:conf/nips/Wei0SBIXCLZ22,DBLP:conf/nips/TianPSJ0HMY24}, self-consistency~\citep{DBLP:conf/iclr/0002WSLCNCZ23}, and multi-step calibration~\citep{DBLP:journals/corr/abs-2503-04625,DBLP:journals/corr/abs-2408-03314}.
Other approaches focus on the token or latent level, including redundancy pruning~\citep{DBLP:journals/corr/abs-2502-12067}, optimization of high-entropy tokens~\citep{DBLP:journals/corr/abs-2506-01939,DBLP:journals/corr/abs-2506-02867}, and latent-space reasoning~\citep{DBLP:journals/corr/abs-2505-16782,DBLP:conf/iclr/00110YW025}.
Recent studies have also explored reasoning mechanisms from internal activations.
\citet{DBLP:journals/corr/abs-2510-13940} improves reasoning via token-level entropy guidance during inference.
\citet{DBLP:conf/emnlp/TangZMZSSW25} identifies reasoning neurons from activation differences for steering, but achieves limited performance gains.
Concurrent work~\citep{DBLP:conf/aaai/GalichinDDRRTO26} leverages Sparse Autoencoders (SAEs) and reasoning-token heuristics to identify interpretable reasoning features for activation steering.
In contrast, AdaRAS directly identifies RCNs without auxiliary modules or token heuristics, and introduces neuron filtering and adaptive intervention to achieve more stable and effective steering.

\paragraph{Mechanistic Interpretability.}  
Recent studies have advanced the understanding of transformer-based language models through techniques such as vocabulary projection~\citep{DBLP:journals/corr/abs-2303-08112}, probing classifiers~\citep{DBLP:journals/coling/Belinkov22,DBLP:journals/tmlr/GurneeNPHTB23}, and sparse autoencoders~\citep{bricken2023monosemanticity,DBLP:conf/acl/Xin0ZWCG0LH025}.
These approaches show that specific neurons could encode semantical features, including truthfulness~\citep{DBLP:conf/iclr/BurnsYKS23,DBLP:journals/corr/abs-2310-06824}, sentiment~\citep{DBLP:journals/corr/abs-2310-15154}, refusal~\citep{DBLP:conf/nips/ArditiOSPPGN24}, and stylistic attributes~\citep{DBLP:conf/iclr/0002HBVPW23,DBLP:conf/nips/StolfoWGBSSN24,DBLP:journals/corr/abs-2310-01405, DBLP:conf/acl/RimskyGSTHT24}.
Building on these insights, a growing body of work explores modifying activations during inference to steer model behavior, achieving regulation of personality traits~\citep{DBLP:journals/corr/abs-2507-21509}, improved instruction following~\citep{DBLP:conf/iclr/StolfoBYHN25, DBLP:conf/icml/LiuY0Z24}, and editing of factual knowledge~\citep{DBLP:conf/acl/SubramaniSP22,hernandez2023inspecting}.
While existing steering methods focus on output-level behaviors, we target trajectory-level reasoning using intervention signals derived from contrastive reasoning traces. 
Crucially, our approach identifies these signals offline, applying them at inference time without task-specific retraining.

\section{Conclusion}
In this work, we propose AdaRAS, an inference-time activation steering framework for improving reasoning reliability in LLMs.
We first demonstrate that reasoning correctness is strongly associated with a small subset of neurons.
AdaRAS identifies these reasoning-critical neurons via mean activation differences between correct and incorrect reasoning trajectories, and further refines them with polarity-based activation filtering.
AdaRAS then applies an adaptive intervention to selectively steer neuron activations during inference.
Extensive experiments across ten reasoning benchmarks show that AdaRAS consistently improves accuracy with only minimal additional computation overhead.
Finally, trajectory-level analyses using latent reasoning features provide mechanistic evidence that AdaRAS stabilizes reasoning paths without affecting semantic modeling level of LLMs.

\section*{Limitations}
Although effective, our work has several limitations.
First, our analysis focuses on the Qwen3 series models and STEM benchmarks. Extending the evaluation to a broader range of model architectures and complex reasoning settings (e.g., spatial or multi-hop reasoning) is necessary to assess the generalizability of our findings for future work.
Second, as with many interpretability studies, our approach relies on constructing contrastive data pairs. This requirement may limit applicability to models at capability extremes, where obtaining paired correct and incorrect reasoning trajectories is non-trivial. Exploring alternative data construction strategies that relax this requirement is an important direction for future work.
Third, to identify RCNs, we focus on neuron-level activations with the mean difference metric. Future work could incorporate other advanced techniques, such as sparse autoencoders or vocabulary projection analyses, to further elucidate the mechanistic underpinnings of RCNs.

\section*{Ethics Statement}
This work investigates inference-time activation steering methods for improving the reasoning reliability of large language models. All experiments are conducted using publicly available models and open-source benchmarks. The study does not involve the collection of personal data, human subjects, or real-world deployment, etc.

\section*{Acknowledgments}
This work was supported by 
the Shandong Provincial Natural Science Foundation (ZR2026QC1180), 
the Natural Science Foundation of China (62372275, 62472261, 62502282, 62506210), 
the foundation of Jinan (JNSX2025004, 202534077), 
the Key R\&D Program of Shandong Province (2025CXGC010112), 
Qingdao Key Laboratory of Trustworthy Artificial Intelligence, Shandong Province Key Laboratory of Network Integration Theory and Technology and Shandong Provincial Laboratory for Humanities and Natural Sciences Collaborative Innovation.

\bibliography{main_ref}

\clearpage
\appendix
\section{Preliminary Study of Probing}
\label{app:preliminary}
We conduct a preliminary probing study on mathematical reasoning benchmarks, including AIME and AMC-12.
First, we construct contrastive datasets consisting of paired correct and incorrect reasoning trajectories for each input as described in \secref{sec:method:parallel}.
Then, using Qwen3 series models (1.7B and 4B), we extract last-token activation values across all layers as input features for probing classifiers.
All samples are randomly partitioned into training and testing sets with a 4:1 ratio.
Subsequently, we train a binary classifier to predict reasoning correctness from last-token activations.
To mitigate the high dimensionality of the activation space, we apply $F$-statistic-based feature selection to identify the most discriminative neurons.

\subsection{Data Construction}
\label{app:probing:data}


We utilize Qwen3-32B to generate contrastive reasoning trajectories.
Specifically, starting from all AIME (24+25) and AMC-12 problems, we sample 8 reasoning traces per question using a generation temperature of 1.0.
We retain only those questions that yield a balanced outcome of exactly 4 correct and 4 incorrect trajectories.
Under this setting, we obtain 15 AIME problems (from an initial 60) and 13 AMC-12 problems (from an initial 104), resulting in 120 and 104 samples, respectively, for probing classifier training.

\subsection{Preprocessing}
\label{app:probing:feature}
For each reasoning trace, we extract last-token activations across all neurons as input features for the probing classifier.
Specifically:
\begin{itemize}
\item \textbf{Source:} Hidden states of the last token in the reasoning path, immediately before final answer generation.
\item \textbf{Position:} Post-activation outputs of the MLP blocks (i.e., after the SwiGLU activation for Qwen3 series model).
\item \textbf{Dimension:} Activations from all transformer layers are concatenated into a single feature vector of dimension $L \times d_{\text{mlp}}$, where $L$ denotes the number of layers.
\item \textbf{Normalization:} Feature values are scaled by dividing by $10 \times \sigma$, where $\sigma$ is the standard deviation for each neuron computed over the training set.
\end{itemize}

\subsection{Probing Classifier}
\label{app:probing:setup}
Given the high dimensionality of the activation space relative to the training sample size, feature selection is critical to mitigate overfitting.
We rank neurons using the ANOVA $F$-statistic (\texttt{sklearn.f\_classif}) based on their linear association with the correctness label, and select the top-$K$ neurons with $K=131{,}072$.
For the probing classifier, we employed a Logistic Regression model with L1 regularization.
The model was trained using 5-fold cross-validation with a stratified 4:1 train-test split. 
The detailed settings are summarized in Table~\ref{tab:probing_hyperparams}.

\begin{table}[ht]
\centering
\caption{Implementation details of probing classifier.}
\label{tab:probing_hyperparams}
\resizebox{0.95\linewidth}{!}{
\begin{tabular}{l|l}
\toprule
\textbf{Setting} & \textbf{Configuration / Value} \\
\midrule
\multicolumn{2}{c}{Feature} \\
\midrule
Source & Last token of reasoning path \\
Position & MLP post-activation output \\
Scope & Global (all layers) \\
Normalization & Scaled by $1 / (10 \times \text{std}_{\text{train}})$ \\
Feature Selection & ANOVA $F$-statistic \\
Dimension of Input & 131,072 \\
\midrule
\multicolumn{2}{c}{Training} \\
\midrule
Architecture & Logistic Regression (\texttt{sklearn}) \\
Regularization & L1 (Lasso) \\
Solver & SAGA \\
Class Weight & Balanced \\
Penalty Strength & Grid search $\in [10^{-4}, 10]$ \\
Max Iterations & 50 \\
Random Seed & 42 \\
Validation & 5-fold cross-validation \\
Metric & AUROC \\
\bottomrule
\end{tabular}
}
\end{table}

\section{Detailed Data Statistics}


\subsection{Data for Adaptive Intervention Module}
\label{app:data:ada_train}
The adaptive intervention module is a binary classifier that predicts whether a given input is likely to fail under the base model and therefore requires intervention.
Training data are constructed by sampling the base model on the training split for each benchmarks.
Samples answered correctly are labeled as negative (i.e., no need for intervention), while incorrect samples are labeled as positive (i.e., intervention required).
Dataset statistics are summarized in Table~\ref{tab:ada_data_stats}.
Notably, for AMC-12 and HumanEval, we evaluate them only in a transfer setting using estimators trained on AIME and MBPP, due to lacking of training split; consequently, they are omitted from Table~\ref{tab:ada_data_stats}.

\begin{table}[ht]
\centering
\caption{Statistics of the training and validation data for the adaptive intervention module.}
\label{tab:ada_data_stats}
\vspace{2mm}
\resizebox{1.0\linewidth}{!}{
\begin{tabular}{l|c|cc}
\toprule
\multirow{2}{*}{\textbf{Dataset}} & \multirow{2}{*}{\textbf{Total Samples}} & \multicolumn{2}{c}{\textbf{Number of each label (Train / Val)}} \\
& & \textbf{No Intervention} & \textbf{Need Intervention} \\
\midrule
AIME & 828 & 392 / 99 & 270 / 67 \\
MATH & 1,000 & 622 / 270 & 78 / 30 \\
GSM8K & 1,700 & 1,097 / 459 & 103 / 41 \\
MBPP & 163 & 83 / 29 & 37 / 14 \\
\bottomrule
\end{tabular}
}
\end{table}

\subsection{Data for Evaluation}
\label{app:data:test_set}




We evaluate AdaRAS on a diverse set of benchmarks, where the dataset statistics are summarized in Table~\ref{tab:dataset_stats}.
All datasets are publicly available.
Specifically, GSM8K, MATH, MBPP, and HumanEval are distributed under open licenses (i.e., MIT or Apache 2.0).
The AIME and AMC datasets are sourced from public academic benchmarks widely accepted for evaluating mathematical reasoning.
We split the datasets as follows:
\begin{itemize}
\item \textbf{AIME.} A unified set of samples is applied across all AIME datasets (2024, 2025, and Extend).
This set contains 15 held-out samples (7 from AIME-24 and 8 from AIME-25), which are strictly excluded from all evaluation sets.
\item \textbf{AMC-12, HumanEval.} Since these datasets do not provide official training splits, we reserve a small subset from the test split for performing AdaRAS.
For example, we hold out 15 HumanEval samples for RCNs identification and evaluate on the remaining 149 samples.
\item \textbf{Other datasets.} For benchmarks with official training splits (GSM8K, MATH, MBPP), patterns are extracted exclusively from the training set.
\end{itemize}

\begin{table*}[ht]
\centering
\caption{Statistics of datasets. \#Probe indicates the number of samples used for identifying RCNs by AdaRAS.}
\label{tab:dataset_stats}
\vspace{2mm}

\begin{threeparttable}
\begin{tabular}{l c c l}
\toprule
\textbf{Dataset} & \textbf{\#Probe} & \textbf{\#Test} & \textbf{Test split} \\
\midrule
\multicolumn{4}{c}{Mathematic} \\
\midrule
AIME-24\tnote{1} & 7 & 23 & Self-split \\
AIME-25\tnote{2} & 8 & 22 & Self-split \\
AIME-Extend\tnote{3} & 15 & 150 & Official \\
MATH-500\tnote{4} \ ~\citep{DBLP:conf/iclr/LightmanKBEBLLS24} & 15 & 500 & Official \\
GSM8K\tnote{5} \ ~\citep{DBLP:journals/corr/abs-2110-14168}& 15 & 1,319 & Official \\
AMC-12\tnote{6} & 13 & 91 & Self-split \\
\midrule
\multicolumn{4}{c}{Coding} \\
\midrule
HumanEval\tnote{7} \ ~\citep{DBLP:journals/corr/abs-2107-03374} & 15 & 149 & Self-split \\
HumanEval+\tnote{8} \ ~\citep{evalplus} & 15 & 149 & Self-split \\
MBPP\tnote{9} \ ~\citep{DBLP:journals/corr/abs-2108-07732} & 15 & 378 & Official \\
MBPP+\tnote{10} \ ~\citep{evalplus} & 15 & 378 & Official \\
\bottomrule
\end{tabular}

\footnotesize 
\begin{tablenotes}
\item[1] \url{https://huggingface.co/datasets/Maxwell-Jia/AIME_2024}
\item[2] \url{https://huggingface.co/datasets/opencompass/AIME2025}
\item[3] \url{https://www.kaggle.com/datasets/hemishveeraboina/aime-problem-set-1983-2024}
\item[4] \url{https://huggingface.co/datasets/HuggingFaceH4/MATH-500}
\item[5] \url{https://huggingface.co/datasets/openai/gsm8k}
\item[6] \url{https://huggingface.co/datasets/rulins/amc12_22-24}
\item[7] \url{https://huggingface.co/datasets/openai/openai_humaneval}
\item[8] \url{https://huggingface.co/datasets/evalplus/humanevalplus}
\item[9] \url{https://huggingface.co/datasets/google-research-datasets/mbpp}
\item[10] \url{https://huggingface.co/datasets/evalplus/mbppplus}
\end{tablenotes}
\end{threeparttable}
\end{table*}

\section{Details of Implementation}

We use the following hyperparameters for AdaRAS in main experiments:
\begin{itemize}
\item \textbf{Number of top-$K$ neurons.} The number of intervened neurons is fixed to $K=50$.
As shown in Figure~\ref{fig:analysis:heatmap}, these neurons are distributed across layers, with a higher concentration in middle-to-late layers.
\item \textbf{Steering strength ($\alpha$).} The best steering strength coefficient $\alpha$ is searched ranging $[0.1, 0.3]$.
\end{itemize}

All experiments are conducted on 4 NVIDIA A100 GPUs and 8 NVIDIA RTX 3090 GPUs.

\subsection{Adaptive Intervention Module}
We adopt a lightweight attention-based classifier for adaptive intervention module.
Instead of using only the last token, this module take all token activations via an attention pooling mechanism to capture the global reasoning state.
Detailed specifications are provided in Table~\ref{tab:ada_estimator}.

\begin{table}[ht]
\centering
\caption{Architecture details for the adaptive intervention module.}
\label{tab:ada_estimator}
\resizebox{0.95\linewidth}{!}{
\begin{tabular}{l|l}
\toprule
\textbf{Setting} & \textbf{Configuration} \\
\midrule
\multicolumn{2}{c}{Architecture} \\
\midrule
Structure & \makecell[l]{Linear(256, 256) \\ $\to$ ReLU \\ $\to$ Dropout(0.3) \\ $\to$ Linear(256, 1)} \\
\midrule
\multicolumn{2}{c}{Training} \\
\midrule
Feature Selection & ANOVA $F$-statistic \\
Dimension of Input & 256 \\
Optimizer & Adam (LR=$10^{-4}$, Weight Decay=$10^{-5}$) \\
Loss Function & BCEWithLogitsLoss \\
Epochs & 100 (Early Stopping Patience=10) \\
\bottomrule
\end{tabular}
}
\end{table}

\subsection{Evaluation}
\label{app:imp:prompts}
To ensure fair comparison and reproducibility, we use greedy decoding during inference across all benchmarks.

\paragraph{Mathematic Benchmarks.} For the AIME, MATH-500, GSM8K, and AMC-12 datasets, we use the following prompt:

\begin{tcolorbox}[colback=gray!10, colframe=gray!50, boxrule=0.5pt, arc=2mm, left=2mm, right=2mm, top=2mm, bottom=2mm]
\small
\texttt{User: Question: \{question\}} \\
\texttt{Thinking process:} \\
\texttt{Please provide a step-by-step thinking process and put your final answer in \textbackslash boxed\{\}.}
\end{tcolorbox}

\paragraph{Coding Benchmarks.} For the HumanEval and MBPP datasets, we use the following prompt:
\begin{tcolorbox}[colback=gray!10, colframe=gray!50, boxrule=0.5pt, arc=2mm, left=2mm, right=2mm, top=2mm, bottom=2mm]
\small
\texttt{User: \{question\}} 
\end{tcolorbox}

We employed the EvalPlus~\citep{evalplus} library to safely extract and execute code blocks.

\section{Details of Reasoning Features}
\label{app:reasoning_feat}

In Section~\ref{sec:analysis:vis}, we use two trajectory-level metrics, Magnitude ($\bar{\mathcal{M}}$) and Angle ($\bar{\mathcal{A}}$), to analyze the geometric properties of reasoning trajectories in latent space.
As in Figure~\ref{fig:analysis:feats}, each point in the kernel density estimation (KDE) plot corresponds to the aggregated $(\bar{\mathcal{M}}, \bar{\mathcal{A}})$ values of a single complete reasoning trace.
These metrics characterize the curvature and directional consistency of representation evolution across layers.
Hidden states $\mathbf{h} \in \mathbb{R}^{d}$ are extracted from the model for each input, and the analysis is performed exclusively on generated reasoning tokens, excluding the input prompt.
Following~\citet{DBLP:conf/iclr/00110YW025}, the formal definitions are as follows:

\paragraph{Sequence-Averaged Magnitude ($\bar{\mathcal{M}}$).}
This metric measures the ``tortuosity'' of the reasoning trajectory in the activation space. 
It is defined as the ratio of the accumulated layer-wise changes to the net change from the first to the last layer. 
For a single token $t$, the magnitude score $\mathcal{M}_t$ is computed as the $L_2$-norm of the difference vector between adjacent layers, normalized by the global displacement:
\begin{equation}
\mathcal{M}_t = \frac{1}{L} \cdot \frac{\sum_{l=0}^{L-1} \| \mathbf{h}_t^{l+1} - \mathbf{h}_t^l \|_2}{\| \mathbf{h}_t^{L} - \mathbf{h}_t^{0} \|_2 + \epsilon},
\end{equation}
where $\epsilon=10^{-6}$ is a small constant for numerical stability.
The final sequence-averaged metric is obtained by averaging over all generated tokens:
\begin{equation}
\bar{\mathcal{M}} = \frac{1}{T} \sum_{t=1}^{T} \mathcal{M}_t.
\end{equation}
A lower $\bar{\mathcal{M}}$ indicates a straighter, more direct transformation of representations through the network depth, which we associate with more stable reasoning.

\begin{table*}[ht]
\centering
\caption{
Comparison with strong test-time baselines, including $k$-sampling, verifier-based reranking, and few-shot CoT prompting.
}
\begin{tabular}{lcccc}
\toprule
Dataset & $k$-Sampling & Verifier & Few-shot & AdaRAS \\
\midrule
AIME-24    & 47.83 {\scriptsize (maj@4)}  & 30.43 {\scriptsize (Best-of-4)} & 39.13 {\scriptsize (3-shot)} & \textbf{60.87} {\scriptsize (Greedy)} \\
AIME-25    & 50.00 {\scriptsize (maj@4)}  & 40.91 {\scriptsize (Best-of-4)} & 31.82 {\scriptsize (3-shot)} & \textbf{54.55} {\scriptsize (Greedy)} \\
\midrule
HumanEval  & 93.96 {\scriptsize (pass@4)} & 79.19 {\scriptsize (Best-of-4)} & -- & \textbf{96.64} {\scriptsize (pass@4)} \\
HumanEval+ & 87.92 {\scriptsize (pass@4)} & 71.82 {\scriptsize (Best-of-4)} & -- & \textbf{89.26} {\scriptsize (pass@4)} \\
\bottomrule
\end{tabular}
\label{tab:app_strong_baselines}
\end{table*}

\begin{table*}[ht]
\centering
\caption{
Comparison with post-trained reasoning models under sampling-based metrics.
AdaRAS uses greedy decoding on AIME and pass@4 on coding tasks.
}
\begin{tabular}{lccccc}
\toprule
Dataset & CoT & R1-Distill & Nemotron & OpenThinker & AdaRAS \\
\midrule
AIME-24 {\scriptsize (maj@4)}     & 47.83 & 26.09 & 52.17 & 52.17 & \textbf{60.87} \\
AIME-25 {\scriptsize (maj@4)}     & 50.00 & 36.36 & 45.45 & 50.00 & \textbf{54.55} \\
\midrule
HumanEval {\scriptsize (pass@4)}  & 93.96 & 75.17 & 78.52 & 83.89 & \textbf{96.64} \\
HumanEval+ {\scriptsize (pass@4)} & 87.92 & 69.80 & 73.83 & 78.52 & \textbf{89.26} \\
\bottomrule
\end{tabular}
\label{tab:app_post_trained}
\end{table*}

\paragraph{Sequence-Averaged Angle ($\bar{\mathcal{A}}$).}
This metric captures the directional stability of the semantic evolution.
It calculates the angular deviation between adjacent layers relative to the global angular shift.
We first define the cosine similarity $\text{sim}(\mathbf{u}, \mathbf{v}) = \frac{\mathbf{u} \cdot \mathbf{v}}{\|\mathbf{u}\|\|\mathbf{v}\|}$.
The angle score $\mathcal{A}_t$ for token $t$ is defined as:
\begin{equation}
\mathcal{A}_t = \frac{1}{L} \cdot \frac{\sum_{l=0}^{L-1} \arccos \left( \text{sim}(\mathbf{h}_t^l, \mathbf{h}_t^{l+1}) \right)}{\arccos \left( \text{sim}(\mathbf{h}_t^0, \mathbf{h}_t^{L}) \right) + \epsilon}.
\end{equation}
Similarly, the sequence-averaged angle is computed as:
\begin{equation}
\bar{\mathcal{A}} = \frac{1}{T} \sum_{t=1}^{T} \mathcal{A}_t.
\end{equation}
Larger values of $\bar{\mathcal{M}}$ or $\bar{\mathcal{A}}$ imply that the reasoning process undergoes significant fluctuations or detours in the latent space.

\section{Additional Experiments}
\label{app:additional}
\subsection{Stronger Test-Time Baselines}
To further examine whether the gains of AdaRAS can be matched by stronger test-time baselines, we compare AdaRAS with three representative alternatives: 
(1) sampling-based methods, including majority voting on mathematical reasoning tasks and pass@4 on coding tasks; 
(2) verifier-based reranking, where a trained verifier selects the best response from multiple generations; and 
(3) few-shot chain-of-thought prompting, where demonstration examples are provided in the prompt.
These baselines increase the inference budget or provide additional test-time supervision, and therefore serve as strong controls for evaluating whether AdaRAS improves reasoning beyond simple statistical scaling or prompt augmentation.

As shown in Table~\ref{tab:app_strong_baselines}, AdaRAS consistently outperforms these alternatives. 
On AIME, AdaRAS uses only single-pass greedy decoding, yet surpasses majority voting with four sampled generations. 
This indicates that its improvement cannot be explained by simple statistical scaling. 
AdaRAS also outperforms verifier reranking, suggesting that directly steering reasoning-critical activations is more effective than selecting among completed trajectories. 
In addition, 3-shot CoT prompting fails to close the gap, showing that the identified reasoning-critical neurons provide transferable activation-level guidance rather than merely encoding prompt demonstrations.

We also compare AdaRAS with post-trained reasoning models under sampling-based metrics. 
As shown in Table~\ref{tab:app_post_trained}, AdaRAS achieves the best performance across all mathematical reasoning and coding benchmarks. 
Notably, on AIME benchmarks, AdaRAS uses single-pass greedy decoding, while the compared baselines are evaluated with majority voting over four sampled generations. 
These results show that AdaRAS provides a compute-efficient and effective test-time enhancement compared with repeated sampling, verifier reranking, few-shot prompting, and post-trained reasoning variants.

\begin{table}[t]
\centering
\caption{
Scalability analysis on Qwen3-4B-Thinking-2507 and Qwen3-14B.
All results are evaluated under our Greedy-CoT setting.
}
\small
\begin{tabular}{lcccc}
\toprule
Dataset & 4B-2507 & +Ours & 14B & +Ours \\
\midrule
AIME-24    & 56.52 & \textbf{65.22} & \textbf{60.87} & \textbf{60.87} \\
AIME-25    & 63.64 & \textbf{77.27} & 72.73 & \textbf{77.27} \\
HumanEval  & 97.99 & \textbf{99.33} & -- & -- \\
HumanEval+ & 92.62 & \textbf{94.63} & -- & -- \\
\bottomrule
\end{tabular}
\label{tab:app_scalability}
\end{table}

\begin{table*}[ht]
\centering
\caption{
Linguistic feature analysis of generated responses.
AdaRAS does not consistently increase response length or structural complexity.
}
\resizebox{\linewidth}{!}{
\begin{tabular}{lcccccccc}
\toprule
\multirow{2}{*}{Dataset} 
& \multicolumn{2}{c}{\#Char} 
& \multicolumn{2}{c}{\#Words} 
& \multicolumn{2}{c}{\#Sentences} 
& \multicolumn{2}{c}{Dep. Depth} \\
\cmidrule(lr){2-3}
\cmidrule(lr){4-5}
\cmidrule(lr){6-7}
\cmidrule(lr){8-9}
& CoT & AdaRAS & CoT & AdaRAS & CoT & AdaRAS & CoT & AdaRAS \\
\midrule
AIME-Extend & 36616.7 & 35966.5 & 7555.6 & 7379.4 & 557.3 & 573.5 & 4.67 & 4.62 \\
Math-500    & 11664.4 & 11581.7 & 2407.4 & 2380.6 & 224.6 & 219.5 & 4.38 & 4.36 \\
GSM8K       & 4765.3  & 4585.2  & 939.3  & 906.1  & 94.4  & 90.9  & 4.34 & 4.34 \\
AMC-12      & 24386.8 & 22399.2 & 5059.1 & 4656.6 & 462.8 & 413.7 & 4.36 & 4.37 \\
\midrule
HumanEval   & 10091.0 & 11192.5 & 1881.7 & 2104.0 & 203.3 & 222.4 & 4.01 & 4.06 \\
MBPP        & 8456.7  & 8118.2  & 1546.7 & 1492.3 & 149.4 & 144.5 & 4.25 & 4.23 \\
\bottomrule
\end{tabular}
}
\label{tab:app_linguistic_features}
\end{table*}

\begin{table}[t]
\centering
\small
\begin{tabular}{lccc}
\toprule
Dataset & Qwen3-4B & +AdaRAS & +AdaRAS \\
        & Sampling & Sampling & Greedy \\
\midrule
AIME-24 & $60.87 \pm 0.00$ & $\textbf{61.96} \pm 2.18$ & \textbf{65.22} \\
AIME-25 & $75.00 \pm 2.62$ & $\textbf{76.14} \pm 1.97$ & \textbf{77.27} \\
\bottomrule
\end{tabular}
\caption{
Reproduction under the official Qwen3 sampling setting.
Sampling results are averaged over four independent runs with different random seeds.
}
\label{tab:app_sampling_reproduction}
\end{table}

\subsection{Model Scalability}

We further evaluate AdaRAS on Qwen3-4B-Thinking-2507 and Qwen3-14B to examine whether the identified reasoning-critical neurons remain effective across model variants and scales. As shown in Table~\ref{tab:app_scalability}, AdaRAS brings consistent improvements on the reasoning-tuned Qwen3-4B-Thinking-2507 model and further improves Qwen3-14B on AIME-25. These results suggest that AdaRAS is not limited to a single base model, and can transfer to both reasoning-tuned and larger models.
For reproducibility, we rerun Qwen3-4B-Thinking-2507 under the same Greedy-CoT evaluation protocol used in this paper. The obtained scores may differ from previously reported results due to differences in decoding and prompting setups.

To account for sampling randomness, we further evaluate AdaRAS under the official Qwen3 sampling setting over four independent runs with different random seeds. 
As shown in Table~\ref{tab:app_sampling_reproduction}, AdaRAS is evaluated with the same decoding configuration and consistently improves over the sampling baseline. Notably, AdaRAS achieves even stronger performance under single-pass greedy decoding, suggesting that its gains do not rely on sampling randomness.

\begin{table}[ht]
\centering
\caption{
Computational cost analysis.
Training time is the offline cost for fitting the gating predictor, while inference overhead is measured per sample.
}
\begin{tabular}{lcc}
\toprule
Dataset & Train (s) & Infer (ms) \\
\midrule
AIME-Extend & 1103.36 & 89.1 \\
Math-500    & 1047.86 & 90.7 \\
GSM8K       & 1674.78 & 108.8 \\
AMC-12      & 35.68   & 91.1 \\
\midrule
HumanEval   & --      & 88.5 \\
MBPP        & 81.84   & 88.0 \\
\bottomrule
\end{tabular}
\label{tab:app_compute_cost}
\end{table}

\subsection{Efficiency Analysis}

\paragraph{Reasoning Length.} 
We analyze whether AdaRAS improves reasoning by simply changing response format or increasing generation length. 
Table~\ref{tab:app_linguistic_features} reports several linguistic features of the generated responses. 
Across different reasoning and coding benchmarks, AdaRAS shows no consistent increase in response length, number of words, number of sentences, or dependency tree depth. 
This suggests that the performance gains do not come from superficial formatting changes or longer reasoning traces.

\paragraph{Computational Cost.} 
We also report the computational cost of our AdaRAS in Table~\ref{tab:app_compute_cost}. 
The gating predictor is trained offline and requires only a small one-time cost. 
During inference, AdaRAS introduces only about $0.09$--$0.11$ seconds of overhead per sample, which is substantially cheaper than sampling-based methods requiring multiple generations.
For HumanEval, we use the predictor trained on MBPP because HumanEval does not provide a training split.
Overall, AdaRAS improves reasoning reliability without substantially increasing output length or inference latency.

\begin{table*}[t]
\centering
\caption{
Impact of prompt variations.
AdaRAS consistently improves over CoT with and without the ``Let's think step by step'' trigger phrase.
}
\resizebox{\linewidth}{!}{
\begin{tabular}{lcccccccc}
\toprule
\multirow{2}{*}{Setting}
& \multicolumn{2}{c}{AIME-24}
& \multicolumn{2}{c}{AIME-25}
& \multicolumn{2}{c}{HumanEval}
& \multicolumn{2}{c}{HumanEval+} \\
\cmidrule(lr){2-3}
\cmidrule(lr){4-5}
\cmidrule(lr){6-7}
\cmidrule(lr){8-9}
& CoT & AdaRAS & CoT & AdaRAS & CoT & AdaRAS & CoT & AdaRAS \\
\midrule
w/ ``let's think''  & 47.83 & \textbf{60.87} & 40.91 & \textbf{54.55} & 74.50 & \textbf{79.19} & 67.79 & \textbf{71.14} \\
w/o ``let's think'' & 30.43 & \textbf{43.48} & 50.00 & \textbf{50.00} & 77.18 & \textbf{79.19} & 69.80 & \textbf{73.15} \\
\bottomrule
\end{tabular}
}
\label{tab:app_prompt_robustness}
\end{table*}

\subsection{Prompt Stability}

We evaluate the robustness of AdaRAS to prompt variations by comparing performance with and without the ``Let's think step by step'' trigger phrase. As shown in 
Table~\ref{tab:app_prompt_robustness}, AdaRAS consistently outperforms CoT under both prompting settings. On AIME-24, CoT degrades substantially without the reasoning trigger, while AdaRAS maintains a clear advantage. 
On coding tasks, where verbose reasoning prompt may sometimes hurts performance, AdaRAS still improves over the corresponding baseline. 
These results indicate that AdaRAS is less sensitive to prompt selection and provides more stable test-time reasoning enhancement.

\subsection{Additional Ablations}
\label{app:additional:ablation}

\begin{table}[t]
\centering
\caption{
Ablation of the adaptive gating component on the probing baseline.
Probing+ adds the same adaptive gating component to Probing, while AdaRAS further uses RCN selection.
}
\resizebox{\linewidth}{!}{
\begin{tabular}{lccc}
\toprule
Dataset & Probing & Probing+ & AdaRAS \\
\midrule
AIME-24     & 43.48 & 52.17 & \textbf{60.87} \\
AIME-25     & 50.00 & 40.91 & \textbf{54.55} \\
AIME-Extend & 48.00 & 49.33 & \textbf{52.67} \\
Math-500    & 85.40 & 84.40 & \textbf{86.40} \\
GSM8K       & 88.32 & 89.01 & \textbf{89.08} \\
AMC-12      & \textbf{73.63} & 72.53 & 70.33 \\
\midrule
HumanEval   & 77.85 & \textbf{79.19} & \textbf{79.19} \\
HumanEval+  & 72.48 & 71.81 & \textbf{73.15} \\
MBPP        & 70.37 & 70.90 & \textbf{72.22} \\
MBPP+       & 59.52 & 59.79 & \textbf{60.58} \\
\bottomrule
\end{tabular}
}
\label{tab:app_probe_predict}
\end{table}

\subsubsection{Adaptive Intervention on Probing Baseline}
We further isolate the effect of the adaptive gating component by adding it to the vanilla probing baseline, denoted as Probe+Predict. As shown in Table~\ref{tab:app_probe_predict}, adaptive gating brings limited improvements over Probe on some datasets, but it does not consistently improve the baseline and remains clearly weaker than AdaRAS. This indicates that the adaptive intervention module alone is not the main source of the gains. The proposed Reasoning-Critical Neuron (RCN) selection strategy is necessary for effective and stable activation steering.

\begin{table}[ht]
\centering
\caption{
Effectiveness of polarity-based filtering.
``Magnitude Only'' selects top-50 neurons solely based on importance scores without polarity filtering.
}
\label{tab:aime_ablation}

\small 
\setlength{\tabcolsep}{1pt} 

\begin{tabular*}{\linewidth}{@{\extracolsep{\fill}} l c c c }
\toprule
\multirow{2.5}{*}{\textbf{Dataset}} & \multicolumn{3}{c}{\textbf{Accuracy (\%)}} \\
\cmidrule{2-4}
& \textbf{Base} & \textbf{Magnitude} & \textbf{Polarity-Aware} \\
&  & \textbf{Only} & \textbf{(Ours)} \\
\midrule
AIME-24      & 47.83 & 56.52 & \textbf{60.87} \\
AIME-25      & 40.91 & 45.45 & \textbf{54.55} \\
AIME-Extended  & 47.33 & 50.67 & \textbf{52.67} \\
\bottomrule
\end{tabular*}
\end{table}

\subsubsection{Effect of Activation Polarity}

In Section~\ref{sec:method:steer}, we proposed a polarity-based filtering method: constructing the steering vector by retaining only neurons that exhibit discriminative sign flips between correct and incorrect traces.
To verify the necessity of this strategy, we conducted an additional ablation study on the AIME benchmarks.
We compare our method against a baseline selection strategy (``Magnitude Only'') that selects the Top-$K$ neurons solely based on the absolute importance score $|S(l,i)|$, without enforcing the polarity constraint.

Table~\ref{tab:aime_ablation} presents the results.
While selecting neurons based on magnitude alone yields performance gains over the base model (e.g., improving AIME-24 from 47.83\% to 56.52\%), introducing the polarity-aware filter significantly amplifies these gains.
Specifically, our method achieves 60.87\% on AIME-24 and 54.55\% on AIME-25, consistently outperforming the magnitude-based selection.
These results empirically confirm that neurons exhibiting contradictory activation patterns (sign flips) are important for reasoning validity, justifying our construction of $\bm S'_l$.

\subsection{Reasoning Trajectories in Latent Space}
\label{app:analysis:trajectory}

To further validate the motivation that correct and incorrect reasoning exhibit distinct activation patterns, we visualize the evolution of latent representations.
We use the weights of the linear probing classifier trained in \secref{sec:pre:probing} to define a reference direction.
Specifically, we project the layer-wise hidden states of reasoning traces onto the direction defined by the top-$K$ most discriminative neurons identified by the probe.

\begin{figure}[t]
\centering
\includegraphics[width=0.95\linewidth]{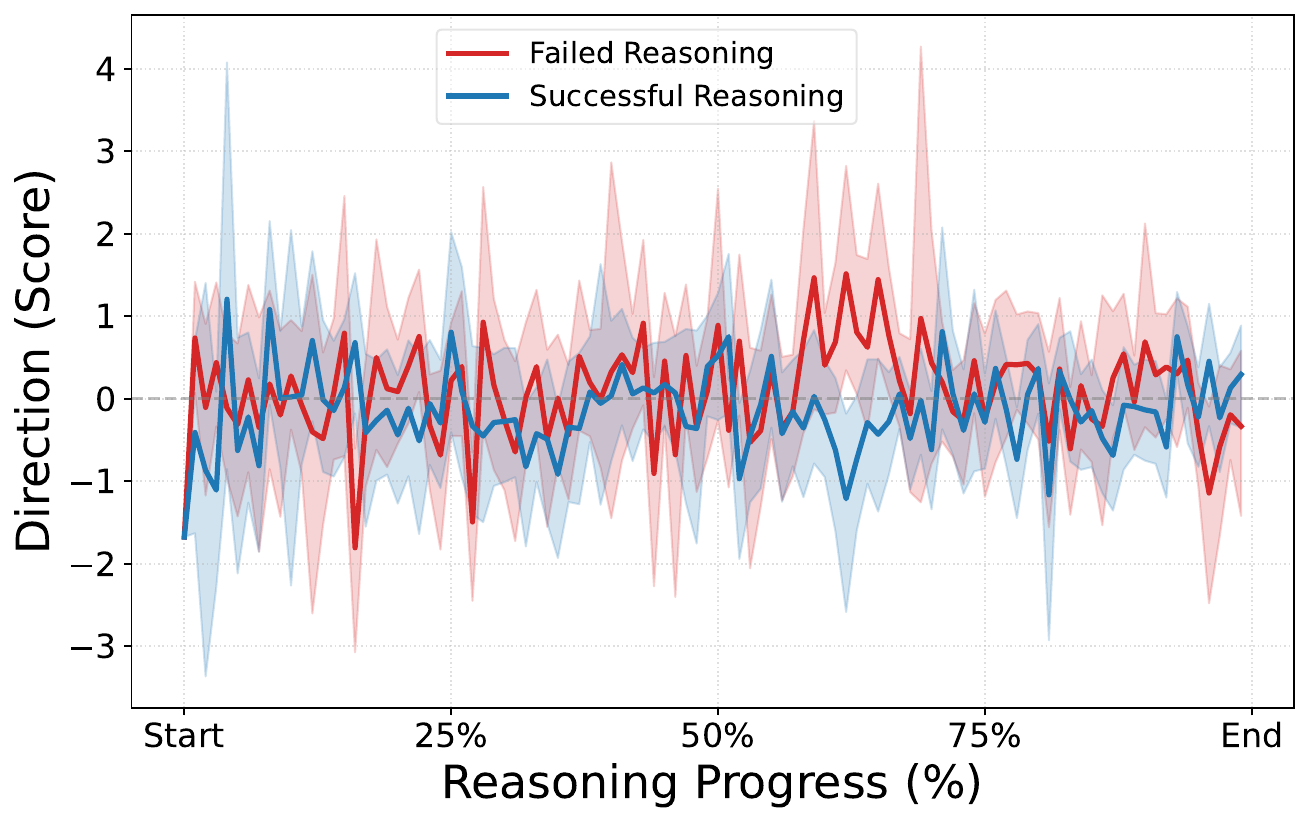}
\caption{
Projection of reasoning states onto the probing classifier learned space.
\textcolor{blue}{Successful (Blue)} and \textcolor{red}{failed (Red)} reasoning processes occupy distinguishable regions in the latent activation space, particularly in the middle and later stages of reasoning.
}
\label{fig:trajectory_vis}
\end{figure}

Figure~\ref{fig:trajectory_vis} illustrates the aggregated trajectories for successful and failed reasoning samples on the AIME dataset.
Although the two curves may overlap locally at certain reasoning stages, they exhibit noticeably different overall trends, particularly in the middle and later stages of reasoning.
Successful traces generally maintain higher projection scores and more stable trajectories, whereas failed traces show larger fluctuations and progressive downward drifts.
These observations suggest that successful and failed reasoning processes occupy distinguishable regions in the latent activation space, supporting the feasibility of extracting reasoning-critical signals from activation patterns.

\section{Case Study}

To intuitively understand how AdaRAS corrects reasoning errors, we present three randomly selected examples from the AIME dataset. 
In each case, we compare the reasoning trace generated by the base model (which fails) against the trace generated with our sparse activation steering.
Blue text highlights the \textcolor{blue}{correct logical steps} induced by AdaRAS, while red text indicates the \textcolor{red}{errors or hallucinations} in the base model's reasoning.

\subsection*{Case 1: Correcting Domain Constraints in Geometry}

\paragraph{Problem:} Let $A, B, C, D$ be points on the hyperbola $\frac{x^2}{20}- \frac{y^2}{24} = 1$ such that $ABCD$ is a rhombus whose diagonals intersect at the origin. Find the greatest real number that is less than $BD^2$ for all such rhombi.

\begin{tcolorbox}[colback=gray!5, colframe=gray!50, title=\textbf{Comparison of Reasoning Traces}]
\small
\textbf{Base Model (Failed):} \\
... We derive the expression for the diagonal squared as $BD^2 = 80 + \frac{22}{3}d^2$. 
\textcolor{red}{Since $d^2 \ge 0$, the expression is always greater than 80. As $d \to 0$, $BD^2$ approaches 80.} 
Therefore, the greatest real number less than $BD^2$ is $\boxed{80}$.

\hrulefill

\textbf{AdaRAS (Success):} \\
... We derive the expression $BD^2 = 80 + \frac{22}{3}d^2$. 
\textcolor{blue}{However, the vertices must lie on the hyperbola. Solving the system for the rhombus vertices $(c,0)$ and $(0,d)$ yields the constraint that $d^2$ cannot be zero. Specifically, the minimum value is found to be $d^2 = \frac{600}{11}$.}
Substituting this into the expression: $BD^2 = 80 + \frac{22}{3} \cdot \frac{600}{11} = 480$.
Thus, the greatest real number is $\boxed{480}$.
\end{tcolorbox}

\paragraph{Analysis:}
In this geometry problem, the Base model correctly derives the algebraic formula but fails to check the \textit{geometric domain constraints} (it assumes $d$ can be arbitrarily small). 
AdaRAS steers the model to verify the existence condition of the rhombus, identifying the lower bound of $d^2$ implied by the hyperbola's equation. 
This illustrates that our steering vector encodes signals related to ``constraint satisfaction.''

\subsection*{Case 2: Rectifying Symbolic Manipulation}

\begin{tcolorbox}[colback=gray!5, colframe=gray!50, title=\textbf{Comparison of Reasoning Traces}]
\small
\textbf{Base Model (Failed):} \\
... We simplify the general term. The rational part telescopes to $\frac{31}{13}$. 
For the logarithmic part, we consider $\prod \log_{k+1} k$. 
\textcolor{red}{This product telescopes to $\frac{1}{\log_4 64} = \frac{1}{3}$.}
Multiplying the results yields $\frac{31}{13} \cdot \frac{1}{3} = \frac{31}{39}$. 
Thus $m + n = \boxed{70}$.

\hrulefill

\textbf{AdaRAS (Success):} \\
... We simplify the general term using logarithmic identities: $\frac{\log_k(5^{k^2-1})}{\log_{k+1}(5^{k^2-4})} \to \dots \to \log_k(k+1)$.
We split the product. The rational part is $\frac{31}{13}$.
\textcolor{blue}{The logarithmic part is $\prod_{k=4}^{63} \log_k(k+1)$. Using the change of base formula $\log_k(k+1) = \frac{\ln(k+1)}{\ln k}$, this telescopes to $\frac{\ln 64}{\ln 4} = \log_4 64 = 3$.}
Multiplying the results gives $\frac{31}{13} \cdot 3 = \frac{93}{13}$. 
Summing them yields $\boxed{106}$.
\end{tcolorbox}

\paragraph{Problem:} Evaluate the product: 
$$\prod_{k=4}^{63} \frac{\log_k(5^{k^2-1})}{\log_{k+1}(5^{k^2-4})}.$$
Find $m+n$ where the result is $m/n$.

\paragraph{Analysis:}
Here, the Base model performs a valid rational simplification but hallucinates the result of the logarithmic telescoping product (calculating it as $1/3$ instead of $3$, effectively inverting the base and argument).
AdaRAS corrects this symbolic manipulation error, guiding the model to apply the change-of-base formula correctly.

\subsection*{Case 3: Deriving Correct Geometric Relationships}
\paragraph{Problem:} Circle $\omega_1$ (radius 6) is internally tangent to $\omega_2$ (radius 15). A rectangle $EFGH$ is inscribed in $\omega_1$ such that specific area conditions are met. Find the area of $EFGH$.

\begin{tcolorbox}[colback=gray!5, colframe=gray!50, title=\textbf{Comparison of Reasoning Traces}]
\small
\textbf{Base Model (Failed):} \\
... Let the vertices be $(a,b)$. Equating the areas of $\triangle DGF$ and $\triangle CHG$ yields the condition $|a(12 - b)| = |b(24 + a)|$.
\textcolor{red}{Simplifying this leads to the relationship $b = \frac{6a}{a + 12}$.}
Substituting this into the circle equation and solving... yields an area of 48.
Final Answer: $\boxed{49}$.

\hrulefill

\textbf{AdaRAS (Success):} \\
... Let the rectangle dimensions be determined by coordinate $y$ and width $m$.
Equating the areas of $\triangle DGF$ and $\triangle CHG$ leads to the equation $12m - my = 48y - my$.
\textcolor{blue}{The term $-my$ cancels out, simplifying to $12m = 48y$, which implies $m = 4y$.}
Substituting this into the circle constraint $(m/2)^2 + y^2 = 36 \implies 5y^2 = 36$.
The area is $m \cdot 2y = 8y^2 = \frac{288}{5}$.
Final Answer: $\boxed{293}$.
\end{tcolorbox}

\paragraph{Analysis:}
The Base model sets up the coordinate geometry correctly but fails at the algebraic simplification step, hallucinating a complex rational relationship ($b = \frac{6a}{a + 12}$) that does not follow from the area equality.
AdaRAS successfully recovers the linear relationship ($m=4y$) by correctly canceling terms during the derivation, ensuring the subsequent area calculation is based on valid premises.

\end{document}